\documentclass[10pt,twocolumn,letterpaper]{article}

\usepackage{cvpr}
\usepackage{times}
\usepackage{epsfig}
\usepackage{graphicx}
\usepackage{amsmath}
\usepackage{amssymb}
\usepackage{enumitem}
\usepackage{gensymb}%

\newcommand{\FigDir}{./figs}

\newcommand{\mS}{{\bf S}}
\newcommand{\mT}{{\bf T}}
\newcommand{\mx}{{\bf x}}
\newcommand{\my}{{\bf y}}

\newcommand{\lc}{\left ( }
\newcommand{\rc}{\right ) }
\newcommand{\lsq}{\left [ }
\newcommand{\rsq}{\right ] } 
\usepackage[pagebackref=true,breaklinks=true,letterpaper=true,colorlinks,bookmarks=false]{hyperref}

\cvprfinalcopy %

\def\cvprPaperID{4235} %

\ifcvprfinal\fi
\begin{document}

\title{DeepMapping: Unsupervised Map Estimation From Multiple Point Clouds}

\author{
 {Li Ding\thanks{This work was partially done while the authors were with MERL. And Chen Feng is the corresponding author.} \ \footnotemark[2]{}}\\
  {\tt\small l.ding@rochester.edu}
 \and
 {Chen Feng\footnotemark[1]{} \ \footnotemark[3]{} \ \footnotemark[4]{}}\\
  {\tt\small cfeng@nyu.edu}
 \and
 \normalsize{\textsuperscript{$\dagger$}University of Rochester \quad \textsuperscript{$\ddagger$}NYU Tandon School of Engineering \quad \textsuperscript{$\mathsection$}Mitsubishi Electric Research Laboratories (MERL)}
 \vspace{-5ex}
}

\maketitle
\thispagestyle{empty}

\begin{abstract}
\vspace*{-3mm}
We propose DeepMapping, a novel registration framework using deep neural networks (DNNs) as auxiliary functions to align multiple point clouds from scratch to a globally consistent frame. We use DNNs to model the highly non-convex mapping process that traditionally involves hand-crafted data association, sensor pose initialization, and global refinement. Our key novelty is that ``training'' these DNNs with properly defined unsupervised losses is equivalent to solving the underlying registration problem, but less sensitive to good initialization than ICP. Our framework contains two DNNs: a localization network that estimates the poses for input point clouds, and a map network that models the scene structure by estimating the occupancy status of global coordinates. This allows us to convert the registration problem to a binary occupancy classification, which can be solved efficiently using gradient-based optimization. We further show that DeepMapping can be readily extended to address the problem of Lidar SLAM by imposing geometric constraints between consecutive point clouds. Experiments are conducted on both simulated and real datasets. Qualitative and quantitative comparisons demonstrate that DeepMapping often enables more robust and accurate global registration of multiple point clouds than existing techniques. Our code is available at \url{https://ai4ce.github.io/DeepMapping/}.
\vspace*{-4mm}
\end{abstract}

\section{Introduction}
\vspace*{-2mm}
\label{sec:intro}

\begin{figure}[t]
    \centering
    \includegraphics[width=1\linewidth]{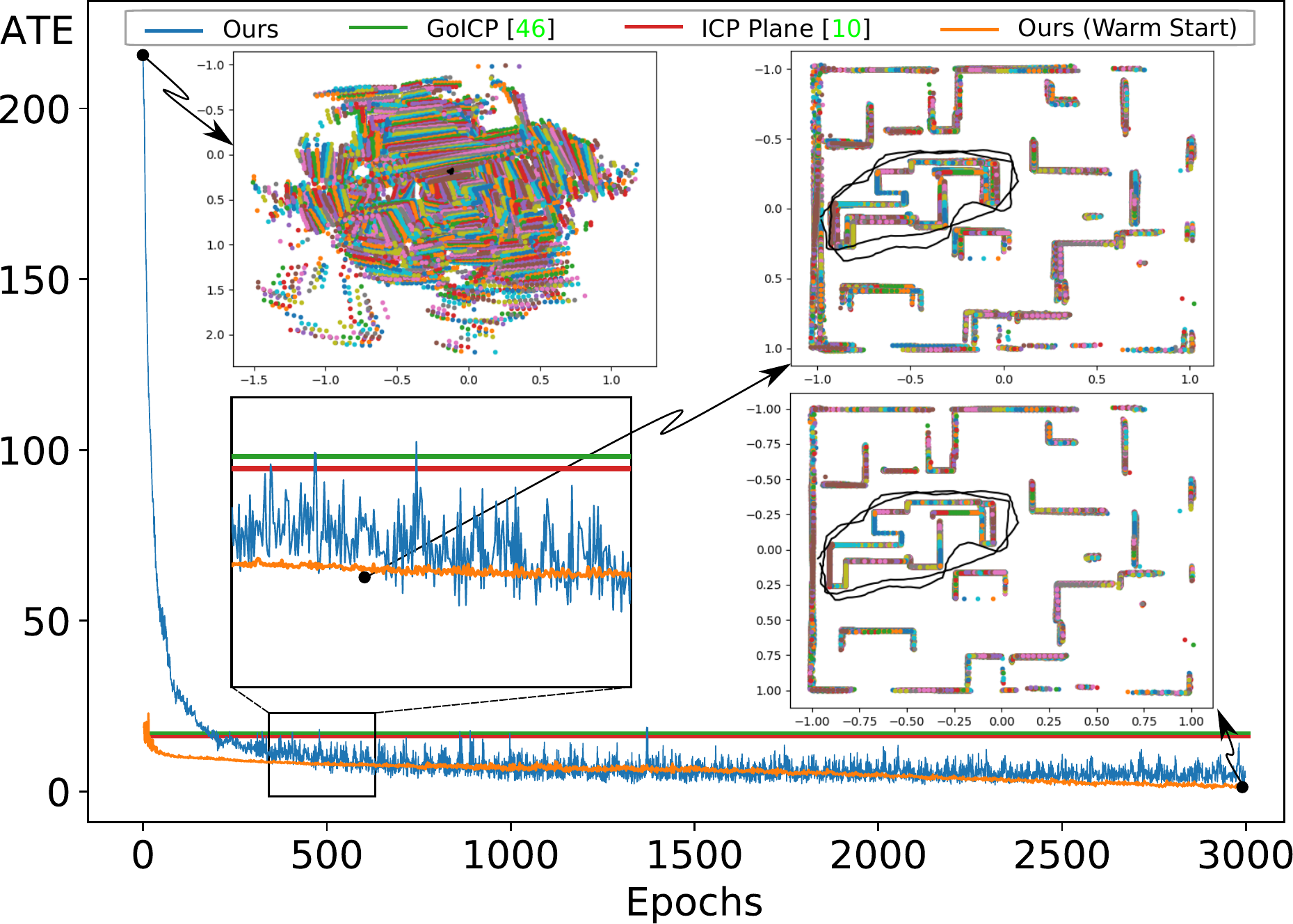}
    \caption{DeepMapping achieves better registration quality than other baselines on an example dataset. Best viewed in color.} %
    \label{fig:1}
\end{figure}
Advances in deep learning have led to many state-of-the-art methods for semantic computer vision tasks. Despite those successes, their compelling improvements on the geometric aspects of computer vision are yet to be fully demonstrated (especially for registration and mapping). This is perhaps because powerful deep semantic representations have limitations in accurately estimating and modeling of geometric attributes of the environment. This includes, but is not limited to, estimating camera motions from a sequence of images, or registering multiple point clouds into a complete model. As opposed to semantic attributes of objects/scenes that are often categorical and thus easily described by human language, those geometric attributes are more often continuous and can be better described numerically, such as poses and shapes. These spatial properties and relations between objects play as vital roles as semantic ones in robotics, augmented reality, medical, and other engineering applications.

Several works attempt to integrate deep learning methods into geometric vision problems~\cite{Ummenhofer_DeMoN_CVPR17,Zhou_SFMLearner_CVPR17,Zhou_DeepTAM_ECCV18,
Henriques_MapNet_CVPR18,Guizilini_LearnOccpMap_IJRR18,
Li_DL2DLoopClos_IROS17,Fraccaro_GenTemModel_ICML18}. Methods in~\cite{Kendall_PoseNet_ICCV15,Brahmbhatt_GeoLearnMap_CVPR18,Henriques_MapNet_CVPR18} try to regress camera poses by training a DNN, inside which a map of the environment is implicitly represented. Methods in~\cite{Ummenhofer_DeMoN_CVPR17,Zhou_SFMLearner_CVPR17} propose unsupervised approaches that exploit inherent relationships between depth and motion. Despite different tasks, most approaches follow the same train-and-test pipeline that neural networks are first learned from a set of training data (either supervised or unsupervised), and then evaluated on a testing set, expecting those DNNs to be able to generalize as much as possible to untrained situations.

The essence of our discussion is an open question: Will DNNs generalize well for geometric problems especially for registration and mapping? Semantic tasks can benefit from DNNs because those related problems are defined empirically, and thus modeled and solved statistically. However many geometric problems are defined theoretically, and thus experiential solutions may not be adequate in terms of accuracy. Think of a simple scenario: given two images with overlapping field-of-views (FOV), without careful calculation, how accurate would a normal person be able to tell the Euclidean distance between the two camera centers? One may argue that reasonable accuracy can be achieved given adequate training. But if this means that it requires a large amount of data collection and training at each new location, the efficiency of this solution seems to be debatable.

Here we investigate another possibility of adopting powerful DNNs for the mapping/registration task. What we commonly agree from abundant empirical experiments and some theorems is that DNNs can model many arbitrarily complex mappings, and can be efficiently optimized through gradient-based methods, at least for categorical classification problems. This leads to our key idea in this paper: we convert the conventionally hand-engineered mapping/registration processes into DNNs, and solve them as if we are ``training'' them, although we do not necessarily expect the trained DNNs to generalize to other scenes. To make this meaningful, unlike the supervised training in~\cite{Kendall_PoseNet_ICCV15,Brahmbhatt_GeoLearnMap_CVPR18,Henriques_MapNet_CVPR18}, we need to properly define unsupervised loss functions that reflect the registration quality. Our exploration towards this line of thought shows promising results in our experiments, as shown in Figure~\ref{fig:1}. We summarize our contributions as follows:
\begin{itemize}[noitemsep,nolistsep]
    \item We propose DeepMapping to solve the point cloud mapping/registration problem as unsupervised end-to-end ``training'' of two DNNs, which is easier for parallel implementation compared to conventional methods requiring hand-crafted features and data associations.
    \item We convert this continuous regression problem to binary classification without sacrificing registration accuracy, using the DNNs and unsupervised losses.
	\item We demonstrate experimentally that DeepMapping is less sensitive to pose initialization compared with conventional baselines.
\end{itemize}

\vspace*{-1mm}
\section{Related Work}
\vspace*{-2mm}
\label{sec:relatedwork}
{\bf Pairwise local registration:} the methods for pairwise point cloud registration can be generally categorized into two groups: local vs. global methods. The local methods assume that a coarse initial alignment between two point clouds and iteratively update the transformation to refine the registration. The typical methods that fall into this category are the Iterative Closest Point (ICP) algorithms~\cite{Besl_ICPPoint_PAMI92, Chen_ICPPlane_IVC92, Rusinkiewicz_EffiICP_DDIM01}, probabilistic-based approaches~\cite{Jian_RegMixGaussians_ICCV05, Myronenko_CPD_PAMI10, Danelljan_ProbColorPCReg_CVPR16} that model the point clouds as a probability distribution. The local methods are well-known for requiring a ``warm start", or a good initialization, due to limited convergence range.

{\bf Pairwise global registration:} the global methods~\cite{Yang_GoICP_PAMI16,Aiger_4PCS_ACMG08,Mellado_Super4PCS_CGF14,Zhou_FastGlobal_ECCV16,Lei_FastDescp_TIP17,Elbaz_PCRegAE_CVPR17} do not rely on the ``warm start'' and can be performed on point clouds with arbitrary initial poses. Most global methods extract feature descriptors from two point clouds. These descriptor are used to establish 3D-to-3D correspondences for relative pose estimation. Robust estimations, e.g., RANSAC~\cite{Fischler_RANSAC_CACM81}, are typically applied to handle the mismatches. The feature descriptors are either hand-crafted such as FPFH~\cite{Rusu_FPFH_ICRA09}, SHOT~\cite{Tombari_SHOT_ECCV10}, 3D-SIFT~\cite{Scovanner_3DSIFT_ACMMM07}, NARF~\cite{Steder_NARF_IROSW10}, PFH~\cite{Rusu_PFH_IROS08}, spin images~\cite{Johnson_SpinImg_PAMI99}, or learning-based such as 3DMatch~\cite{Zeng_3DMatch_CVPR17}, PPFNet~\cite{Deng_PPFNet_CVPR18}, and 3DFeatNet~\cite{Yew_3DFeatNet_ECCV18}.

{\bf Multiple registration:} in addition to pairwise registration, several methods have been proposed for multiple point clouds registration~\cite{Theiler_GloTerrGraOpt_JPRS15,Evangelidis_JRMPC_ECCV14,Izadi_KinectFusion_ACMUIST11,Torsello_MultiRegDualQuant_CVPR11,Choi_ReconIndoor_CVPR15}. One approach is to incrementally add new a point cloud to the model registered from all previous ones. The drawback of the incremental registration is the accumulated registration error. This drift can be mitigated by minimizing a global cost function over a graph of all sensor poses~\cite{Choi_ReconIndoor_CVPR15,Theiler_GloTerrGraOpt_JPRS15}.

{\bf Deep learning approaches: } recent works explore the idea of integrating learning-based approaches into mapping and localization problems.
Methods in~\cite{Ummenhofer_DeMoN_CVPR17,Zhou_SFMLearner_CVPR17} propose unsupervised approaches that exploit inherent relationships between depth and motion.
This idea is further explored in~\cite{CodeSLAM_Bloesch_2018_CVPR,Zhou_DeepTAM_ECCV18,Yand_DVSO_ECCV18,Li_DL2DLoopClos_IROS17} using deep learning for visual odometry and SLAM problems.
For instance, CodeSLAM~\cite{CodeSLAM_Bloesch_2018_CVPR} represents the dense geometry using a variational auto-encoder (VAE) for depth that is conditioned on the corresponding intensity image, which is later optimized during bundle adjustment. Differently, DeepMapping does not require any pre-training. \cite{Fraccaro_GenTemModel_ICML18} introduces a generative temporal model with a memory system that allows the agent to memorize the scene representation and to predict its pose in a partially observed environment. Although this method and DeepMapping are both able to determine the sensor pose from the observed data, \cite{Fraccaro_GenTemModel_ICML18} requires a supervised training stage for ``loading'' its memory, while DeepMapping is fully unsupervised and does not follow the aforementioned train-and-test pipeline. Methods in~\cite{Henriques_MapNet_CVPR18,Parisotto_GlbPoseEstAttRNN_CVPR18} use the recurrent neural network (RNN) to model the environment through a sequence of images in a supervised setting. MapNet~\cite{Henriques_MapNet_CVPR18}, for example, develops a RNN for RGB-D SLAM problem where the localization of camera sensor is performed using deep template matching on the discretized spatial domain that has a relatively small resolution. Unlike MapNet, the proposed DeepMapping does not require any partition of the space and is unsupervised.
\begin{figure*}[t]
    \centering
    \includegraphics[width=1\linewidth]{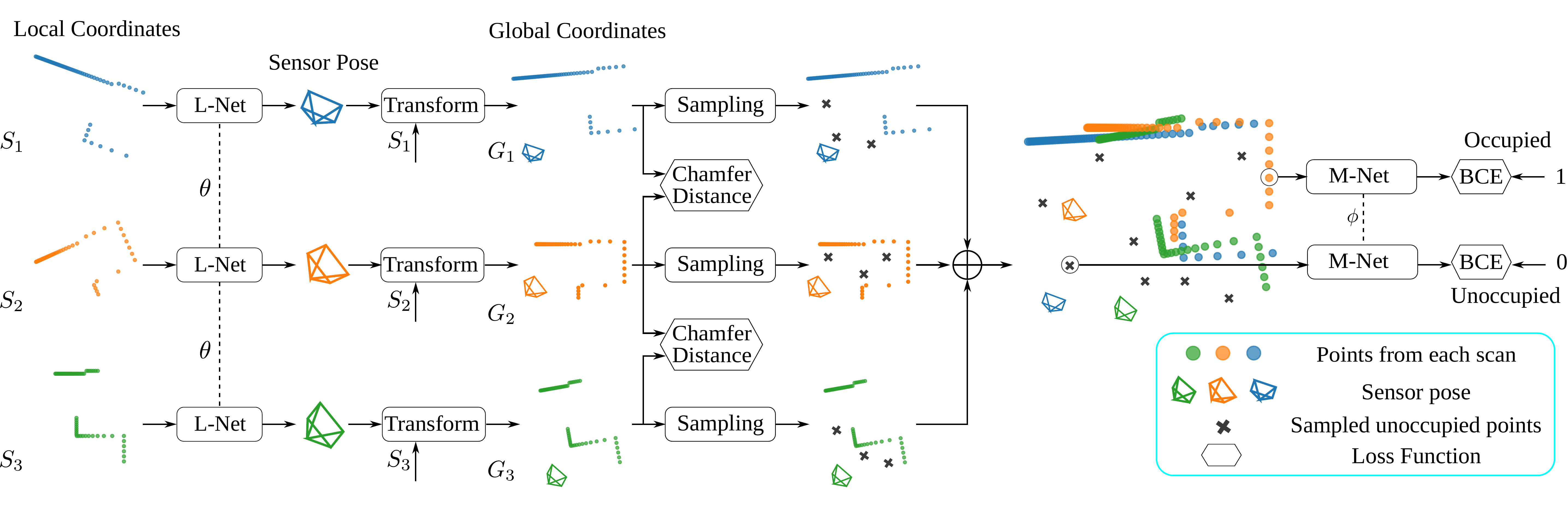}
    \caption{\textbf{DeepMapping pipeline}. Point clouds appear in different colors. Each input point cloud is fed into the shared L-Net to compute transformation parameters that map the input to the global coordinates where both occupied (colored solid circles) and unoccupied (gray cross marks) locations are sampled. The M-Net predicts the occupancy probabilities of the sampled locations. The global occupancy loss is the binary cross entropy (BCE) averaged over all sampled locations. DeepMapping is able to handle temporal information, if available, by integrating Chamfer distance loss between consecutive scans. Best viewed in color.}
    \label{fig:overview}
    \vspace*{-3mm}
\end{figure*}

Other related methods such as~\cite{Kendall_PoseNet_ICCV15,Brahmbhatt_GeoLearnMap_CVPR18} solve camera localization by training DNNs to regress camera poses and test the performance in the same environment as the training images. Related to that is DSAC~\cite{Brachmann_DSAC_CVPR17} as a differentiable alternative to traditional RANSAC for its use in pose estimation DNNs. For place recognition, semantic scene completion is used in~\cite{Schonberger_SemVisLoc_JPRS18} as an auxiliary task for training an image VAE for long-term robustness. The method in~\cite{Guizilini_LearnOccpMap_IJRR18} proposes an unsupervised approach with variational Bayesian convolutional auto-encoder to model structures from point clouds. In DeepMapping, we adopt this idea to model the scene structure but use DNN rather than Bayesian inference. Other prior works include: in~\cite{Eslami_GQN_Science18} the generative query network (GQN) shows the ability to represent the scene from a given viewpoint and rendering it from an unobserved viewpoint in the simple synthetic environments. A neuroscience study~\cite{Banino_VecNaviGridRep_Nature18} uses recurrent neural networks to predict mammalian spatial behavior.

As noted, most approaches follow a train-and-test pipeline. Our approach adopts DNNs but differs from the existing methods in the way that the process of ``training" in DeepMapping is equivalent to solving the point clouds registration and that once trained, we do not expect the DNNs to generalize to other scenes.

\vspace*{-2mm}
\section{Method}
\label{sec:method}
\subsection{Overview}
\label{ssec:overview}
\vspace*{-1mm}
In this section, we describe the proposed DeepMapping that uses DNNs for registering multiple point clouds. Let $\mS=\{S_i\}_{i=1}^{K}$ be the set of $K$ input point clouds in the $D$-dimensional space that are captured by Lidar scanners, and the $i^{th}$ point cloud $S_i$, represented as a $N_i\times D$ matrix, contains $N_i$ points in sensor local frame. Given $K$ point clouds, the goal is to register all point clouds in a common coordinate frame by estimating the sensor poses $\mT = \{T_i\}_{i=1}^{K}$ for each point cloud $S_i$, where $T_i\in SE(D)$.

Conventional methods~\cite{Zhou_FastGlobal_ECCV16,Evangelidis_JRMPC_ECCV14} formulate this as an optimization problem that directly seeks the optimal sensor poses $\mT$ to minimize the loss function
\begin{equation}
\vspace*{-1mm}
\mT^{\star}(\mS) = \operatorname*{arg\,min}_\mT \mathcal{L} \lc \mT, \mS \rc \textrm{,}
\label{eq:trad_loss}
\vspace*{-1mm}
\end{equation}
where $\mathcal{L} \lc \mT, \mS \rc $ is the objective that scores the registration quality. As explained in Section~\ref{sec:intro}, instead of directly optimizing $\mT$, we propose to use a neural network, modeled as an auxiliary function $f_\theta(\mS)$, to estimate $\mT$ for the input point clouds $\mS$ where $\theta$ are the auxiliary variables to be optimized. The pose is then converted to a transformation matrix that maps $S_i$ into its global version $G_i$.

Formally, we re-formulate this registration problem as finding the optimal network parameters that minimize a new objective function
\begin{equation}
(\theta^{\star},\phi^{\star}) = \operatorname*{arg\,min}_{\theta,\phi} \mathcal{L}_\phi \lc f_\theta \lc \mS \rc ,  \mS  \rc \textrm{,}
\label{eq:prop_loss}
\vspace*{-2mm}
\end{equation}
where $\mathcal{L}_\phi$ is an unsupervised learnable objective function which will be explained in Section~\ref{ssec:loss} and~\ref{ssec:slam}.

Figure~\ref{fig:overview} illustrates the pipeline for DeepMapping. At the heart of DeepMapping are two networks, a localization network (L-Net) and an occupancy map network (M-Net), that estimates $T_i$ and measures the registration quality, respectively. The L-Net is a function $f_\theta: S_i \mapsto T_i$ appeared in~\eqref{eq:prop_loss} that estimates the sensor pose of a corresponding point cloud, where the parameters $\theta$ in the L-Net are shared among all point clouds. The point cloud $G_i$ in global coordinates is obtained using the estimated sensor pose. From the transformed point clouds, we first sample both occupied and unoccupied locations. Then the locations of these samples are fed into the M-Net, to evaluate the registration performance of the L-Net. The M-Net is a binary classification network that predicts probabilities of input locations being occupied. We denote M-Net as a function with learnable parameters $\phi$. Those occupancy probabilities are used for computing the unsupervised loss $\mathcal{L}_\phi$ in~\eqref{eq:prop_loss} that measures the global occupancy consistency of the transformed point clouds and thus reflects the registration quality.

One may find that transforming from~\eqref{eq:trad_loss} to~\eqref{eq:prop_loss} could increase the dimensionality/complexity of the problem. In fact, we provide a simple 1D version of this problem conversion and show that using DNNs as auxiliary functions and optimizing them in higher dimensional space using gradient-based methods could enable faster and better convergence than directly optimizing the original problem. Details are included in the supplementary material.

\subsection{DeepMapping Loss}
\label{ssec:loss}
We use the M-Net $m_\phi$ to define the unsupervised loss function $\mathcal{L}_\phi$ that scores the registration quality. The M-Net is a continuous occupancy map $m_\phi: \mathbb{R}^{D} \to \left[ 0,1 \right] $ that maps a global coordinate to the corresponding occupancy probability, where $\phi$ are learnable parameters. If a coordinate in the global frame is associated with a binary occupancy label $y$ indicating whether the location is occupied,
then we can calculate the loss of a global coordinate as the binary cross entropy (BCE) $B$ between the predicted occupancy probability $p$ and the label $y$:
\begin{equation}
B\lsq p, y \rsq = - y \log\lc p \rc - \lc 1-y \rc \log\lc 1-p \rc \textrm{.}
\label{eq:bce}
\vspace*{-1mm}
\end{equation}

The question is how to determine the label $y$.
We approach this question by considering the situation when all point clouds are already globally aligned precisely.
In such a case, it is obvious that all observed/scanned points, in the global frame, should be marked as occupied with the label 1, due to the physical principle of the Lidar scanner.

\begin{figure}[t]
\begin{minipage}[b]{0.49\linewidth}
  \centering
  \includegraphics[width=1\linewidth]{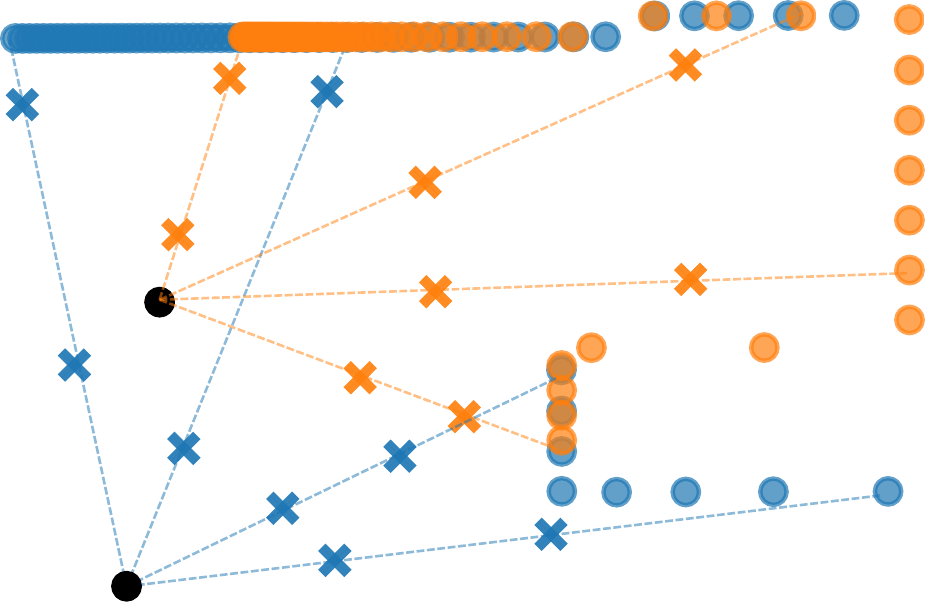}
  \centerline{(a)}
\end{minipage}
\hfill
\begin{minipage}[b]{0.49\linewidth}
  \centering
  \includegraphics[width=1\linewidth]{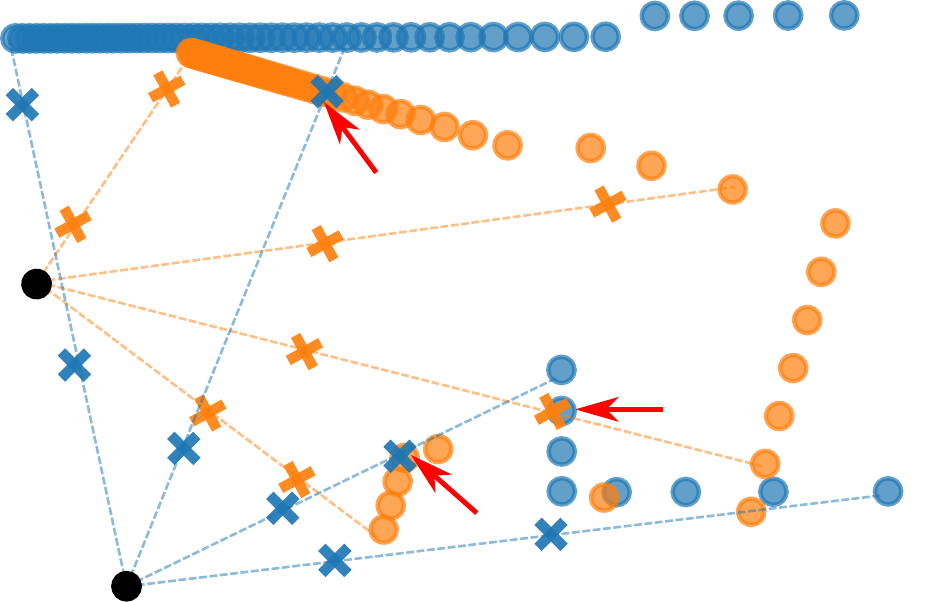}
  \centerline{(b)}
\end{minipage}
\caption{Illustration of sampling methods and self-contradictory occupancy status. Blue and orange circles are two point clouds in the global coordinates and cross marks represent the sampled unoccupied points. (a) and (b) show the correctly aligned and misaligned point clouds, respectively. The red arrows in (b) highlight the points with self-contradictory occupancy status. Best viewed in color.}
\label{fig:sampling} 
\vspace*{-2mm}
\end{figure}

It is also clear now that points lie between the scanner center and any observed points, i.e., on the line of sight, should be marked as unoccupied with label 0. Figure~\ref{fig:sampling} depicts the mechanism to sample unoccupied points for Lidar scanners. The dash lines show the laser beams emitted from the scanner. We denote $s\lc G_i \rc$ as a set of sampled points from $G_i$ that lie on these laser beams, illustrated as cross marks. These points are associated with label $0$ indicating unoccupied locations. %

By combining the binary cross entropy and the sampling function, the loss used in~\eqref{eq:prop_loss} is defined as an average of the binary cross entropy over all locations in all point clouds:
\begin{equation} \label{eq:obj_cls}
\mathcal{L}_{cls} = \frac{1}{K} \sum_{i=1}^{K} B \lsq m_\phi \lc G_i\rc,1 \rsq + B \lsq m_\phi \lc s \lc G_i \rc \rc, 0 \rsq \textrm{,}
\vspace*{-2mm}
\end{equation}
where $G_i$ is a function of L-Net parameters $\theta$, and with a slight abuse of notation $B \lsq m_\phi \lc G_i\rc,1 \rsq$ denotes the average BCE error for all points in point cloud $G_i$, and $B \lsq m_\phi \lc s \lc G_i \rc \rc, 0 \rsq$ means the average BCE error for correspondingly sampled unoccupied locations.

In Figure~\ref{fig:sampling}, we illustrate the intuition behind~\eqref{eq:obj_cls}. The loss function can achieve smaller values if registrations are accurate, as shown in Figure~\ref{fig:sampling} (a). Conversely, since misaligned point clouds will lead to self-contradictory occupancy status, the loss function will be larger due to the difficulties of the M-Net to model such noisy occupancy patterns, as shown in Figure~\ref{fig:sampling} (b). 

Note that the loss in~\eqref{eq:obj_cls} is unsupervised because it relies only on the inherent occupancy status of point clouds rather than any externally labeled ground truth. For minimization, we adopt the gradient-based optimization because the loss function $\mathcal{L}_{cls}$ is differentiable for both $\theta$ and $\phi$. In each forward pass, unoccupied locations are randomly sampled.

It is also worth noting the potentials of M-Net in robotics. Unlike prior works that use discrete occupancy maps, we directly feed the M-Net with the floating number coordinates, resulting in a continuous occupancy map. Therefore, the M-Net offers a new way to represent the environment at an arbitrary scale and resolution.

\subsection{Network Architecture}
\vspace*{-1mm}
\label{ssec:arch}
{\bf L-Net:} the goal of the localization network, L-Net, is to estimate the sensor pose $T_i$ in a global frame. The L-Net consists of a latent feature extraction module followed by a multi-layer perceptron (MLP) that outputs sensor pose $T_i$. This module depends on the format of the input point cloud $S_i$. If $S_i$ is an \textit{organized point cloud} that is generated from range images such as disparity maps or depth images, then $S_i$ is organized as an array of points in which the spatial relationship between adjacent points is preserved. Therefore, we apply a convolutional neural network (CNN) to extract the feature vector of point cloud followed by a global pooling layer to aggregate local features to a global one.
In the case where the inputs are a set of \textit{unorganized point clouds} without any spatial order, we adopt PointNet~\cite{Qi_PointNet_CVPR17} architecture for extracting features from the point cloud. We remove the input and feature transformations appeared in~\cite{Qi_PointNet_CVPR17} and use a shared MLP that maps each $D$-dimensional point coordinate to a high dimensional feature space. A global pooling layer is applied across all points for feature aggregation.%
The extracted latent feature vector is processed with an MLP with the output channels corresponding to the degree of freedom of sensor movement. 

{\bf M-Net:} the occupancy map network, M-Net, is a binary classification network that takes as input a location coordinate in the global space and predicts the corresponding occupancy probability for each input location. The M-Net is an MLP with a $D$-channel input and a 1-channel output from a sigmoid function, shared across all points.

\subsection{Extension to Lidar SLAM}
\vspace*{-1mm}
\label{ssec:slam}
The loss function in~\eqref{eq:obj_cls} treats the input point clouds $\mS$ as an unordered set of scans instead of a temporally ordered sequence. In some applications, the temporal information may be available. For example, Lidar SLAM uses laser scanners to explore the unknown environment and captures an ordered sequence of point clouds at different time $t$. 

Now we extend DeepMapping to exploit such a temporal relationship. The underlying assumption is that the consecutive scans of point clouds are expected to have a reasonable overlapping with each other
, which normally holds in the SLAM settings~\cite{Sturm_TUM_IROS12,Song_Sun3D_CVPR15}. We utilize the geometric constraints among point clouds that are temporally close to each other. Specifically, we adopt the Chamfer distance as the metric that measures the distance between two point clouds $X$ and $Y$ in the global coordinates
 \begin{equation}
    \begin{split}
        d\lc X,Y \rc &= \frac{1}{|X|} \sum_{\mx \in X} \min_{\my \in Y} \| \mx -  \my \|_2   \\
        &+ \frac{1}{|Y|} \sum_{\my \in Y} \min_{\mx \in X} \| \mx -  \my\|_2 \textrm{.}
    \end{split}
   \label{eq:Chamfer}  
   \vspace*{-2mm}
 \end{equation}
The Chamfer distance measures the two-way average distance between each point in one point cloud to its nearest point in the other. The minimization of the Chamfer distance $d\lc G_i,G_j\rc$ results in a pairwise alignment between point clouds $G_i$ and $G_j$. To integrate the Chamfer distance into DeepMapping, we modify the objective in~\eqref{eq:prop_loss} as 
\begin{equation}
(\theta^\star, \phi^\star) = \operatorname*{arg\,min}_{\theta,\phi} \mathcal{L}_{cls} + \lambda \mathcal{L}_{ch} \textrm{,}
\label{eq:opt_cls_ch}
\vspace*{-2mm}
\end{equation}
where $\lambda$ is a hyperparameter to balance the two losses and $\mathcal{L}_{ch}$ is defined as the average Chamfer distance between each point cloud $G_i$ and its temporal neighbors $G_j$
\begin{equation}
\vspace*{-1mm}
    \mathcal{L}_{ch} = \sum_{i=1}^{K}\sum_{j\in\mathcal{N}\lc i \rc} d\lc G_i,G_j \rc \textrm{.}
    \label{eq:obj_ch}
    \vspace*{-1mm}
\end{equation}

\subsection{Warm Start}
\label{ssec:warm_start}
Optimizing~\eqref{eq:opt_cls_ch} with random initialization of network parameters (i.e., ``cold start'') could take a long time to converge, which is undesirable for real-world applications.
Fortunately, DeepMapping allows for seamless integration of a ``warm start'' to reduce the convergence time with improved performance. Instead of starting from scratch, we can first perform a coarse registration of all point clouds using any existing methods, such as incremental ICP, before further refinement by DeepMapping. Figure~\ref{fig:1} shows an example result of registration error versus optimization iterations. With such a warm start, DeepMapping converges much faster and more stable than the one starting from scratch and is more accurate and robust.

\section{Experiments}
\vspace*{-1mm}
\label{sec:exp}
We evaluate DeepMapping on two datasets: a simulated 2D Lidar point cloud dataset and a  real 3D dataset called Active Vision Dataset (AVD)~\cite{Ammirato_AVD_ICRA17}. We implemented DeepMapping with PyTorch~\cite{PytorchWebsite}. The network parameters are optimized using Adam optimizer~\cite{Kingma_Adam_ICLR15} with a learning rate of $0.001$ on an Nvidia TITAN XP. 

For quantitative comparison, we use two metrics: the absolute trajectory error (ATE) and the point distance between a ground truth point cloud and a registered one. The ATE assesses the global consistency of the estimated trajectory, and the point distance is the average distance of corresponding points between the ground truth and the registered point cloud. Since the estimated position can be defined in an arbitrary coordinate, a rigid transformation is determined to map the points in estimated coordinates to the ground truth coordinates using the closed-form solution proposed in~\cite{Horn_ClosedFormTrans_JOSAA87}. The metrics are calculated after such a registration.

\subsection{Experiments on 2D Simulated Point Cloud }
\begin{figure}
\centering
\includegraphics[width=1\linewidth]{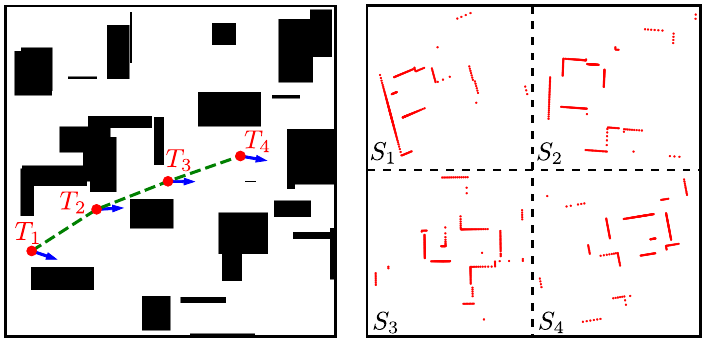}
\caption{Illustration of $4$ scans captured at time $20,70,120,170$. Left: the binary image shows the simulated environment with a size of $1024\times 1024$. The sensor poses are shown in red circles and blue arrows. The green dash line indicates the trajectory of the sensor. Right: observed point clouds captured at corresponding poses. Best viewed in color.}
\vspace*{-2mm}
\label{fig:2d_data}
\end{figure}

\vspace*{-1mm}
\label{ssec:exp_simulated}
{\bf Dataset:} to simulate the 2D Lidar point clouds captured by a virtual Lidar scanner, we create a large environment represented by a $1024\times 1024$ binary image, as shown in Figure~\ref{fig:2d_data}. The white pixels indicate the free space whereas the black ones correspond to obstacles. We assume a moving robot equipped with a horizontal Lidar scanner to explore this environment. We first sample the trajectory of the robot movement that consists of a sequence of poses. The rotation perturbation between two scans are randomly selected from $-10\degree$ to $10\degree$, and the translation perturbation on average is approximately 8.16 pixels. At each pose in the trajectory, the Lidar scanner spreads laser beams over the FOV and the laser beam is reflected back whenever it hits the obstacles within its path. Each observed point is computed as the intersection point between the laser ray and the boundary of obstacles. The scanning procedure yields a set of 2D points in the robot's local coordinate frame. 

In this experiment, we set the FOV of the Lidar scanner to $360 \degree$ and assume an ideal scanner with an unlimited sensing range~\footnote{We tested the robustness of DeepMapping by imposing a limited sensing range, while our resulting performance does not change significantly.}. If no object exists along the laser ray, we return the point where the ray hits the image boundary. We create $3$ binary maps of the virtual environment and sample in total $75$ trajectories. $50$ trajectories have $128$ poses and the remaining 25 trajectories contain 256 poses where point clouds are scanned. Each scan contains 256 points. The transformation $T_i$ is parameterized by a 2D translation vector $\lc t_x,t_y \rc$ and a rotation angle $\alpha$.

{\bf Baseline:} we compare DeepMapping with the following baselines: incremental iterative closest point (ICP) with point-to-point distance metrics~\cite{Besl_ICPPoint_PAMI92} and point-to-plane distance metrics~\cite{Chen_ICPPlane_IVC92}, GoICP~\cite{Yang_GoICP_PAMI16}, and a particle swarm optimization PSO~\cite{PSO_Github}. To justify the advantage of the neural network based optimization, we also test another baseline method, referred to as the direct optimization. Specifically, we remove the L-Net $f_\theta$ and perform the optimization of the same loss function with respect to the sensor poses $\mT$ rather than network parameters $\theta$. 

{\bf Implementation:} the detailed architecture of the L-Net consists of C$\lc 64 \rc$-C$\lc 128 \rc$-C$\lc1024 \rc$-M$\lc 1024 \rc$-FC$\lc 512\rc$-FC$\lc 256\rc$-FC$\lc 3\rc$, where C$\lc n\rc$ denotes 1D atrous convolutions that have kernel size $3$, dilation rate of $2$ and $n$-channel outputs,  M$\lc n \rc$ denotes 1D global max-pooling over $n$ channels, FC$\lc n \rc$ denotes fully-connected layer with $n$-channel output. The M-Net can be described as FC$\lc 64\rc$-FC$\lc 512 \rc$-FC$\lc 512 \rc$-FC$\lc 256 \rc$-FC$\lc 128 \rc$-FC$\lc 1 \rc$. We did not optimize the network architectures. More ablation studies are included in the supplementary material. To sample unoccupied points, we randomly select $19$ points on each laser ray. We run PSO for multiple rounds until they consume the same time as DeepMapping. To compare DeepMapping with the direct optimization, we use the same optimizer and the learning rate. To ensure the same initialization between DeepMapping and the direct optimization, we run DeepMapping with only on forward pass (no back-propagation is performed) and save the output sensor poses. These poses are used as the starting point for the direct optimization. The hyper-parameter $\lambda$ is chosen based on the prior knowledge of the dataset. For the 2D dataset where sequential frames have reasonable overlaps, we assign more weights to Chamfer loss by setting $\lambda$ to 10. The batch size is 128, and the optimization speed is approximately 100 epochs per minute on an Nvidia TITAN XP.

\begin{figure*}[ht]
\centering
\includegraphics[width=1\linewidth]{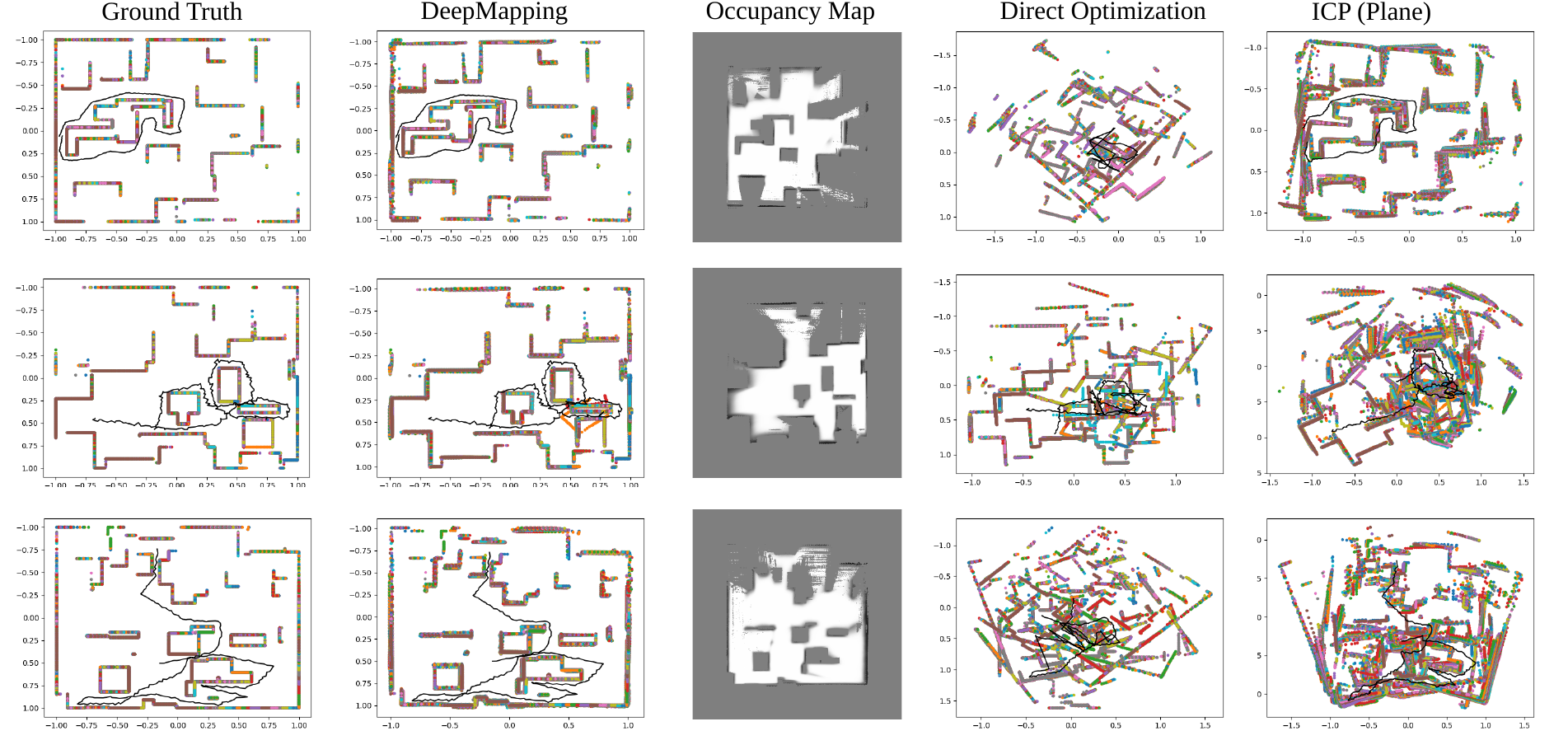}
\caption{Qualitative results from the 2D simulated dataset. The black lines are the sensor trajectories. The third column shows the occupancy maps for each trajectory that are estimated by the M-Net. The black, white, and gray pixels show the occupied, unoccupied, and unexplored locations, respectively. While the results of each trajectory are defined in arbitrary coordinate systems, we aligned them with the ground truth coordinates for a clear comparison. More visualization results are included in supplementary material. Best viewed in color.}
\label{fig:2d_qual}
\end{figure*}

{\bf Results:} 
Figure~\ref{fig:2d_qual} shows the qualitative comparison of 3 trajectories simulated from the dataset.  As shown, the baseline method, direct optimization on sensor pose, fails to find the transformations, leading to globally inaccurate registration. While the incremental ICP algorithms are able to procedure a noisy registration, errors of the incremental ICP algorithms accumulate over frames. As opposed to the baselines, the proposed DeepMapping accurately aligns multiple point clouds by utilizing neural networks for the optimization. Once the registration is converged, we feed all points (both occupied and unoccupied) in the global coordinates into the M-Net to estimated the occupancy probability, which is then converted to an image of occupancy map shown in the third column in Figure~\ref{fig:2d_qual}. The unexplored regions are shown in gray. The estimated occupancy map agrees with registered point clouds.

\begin{figure}[ht]
\begin{minipage}[b]{0.49\linewidth}
  \centering
  \includegraphics[width=1\linewidth]{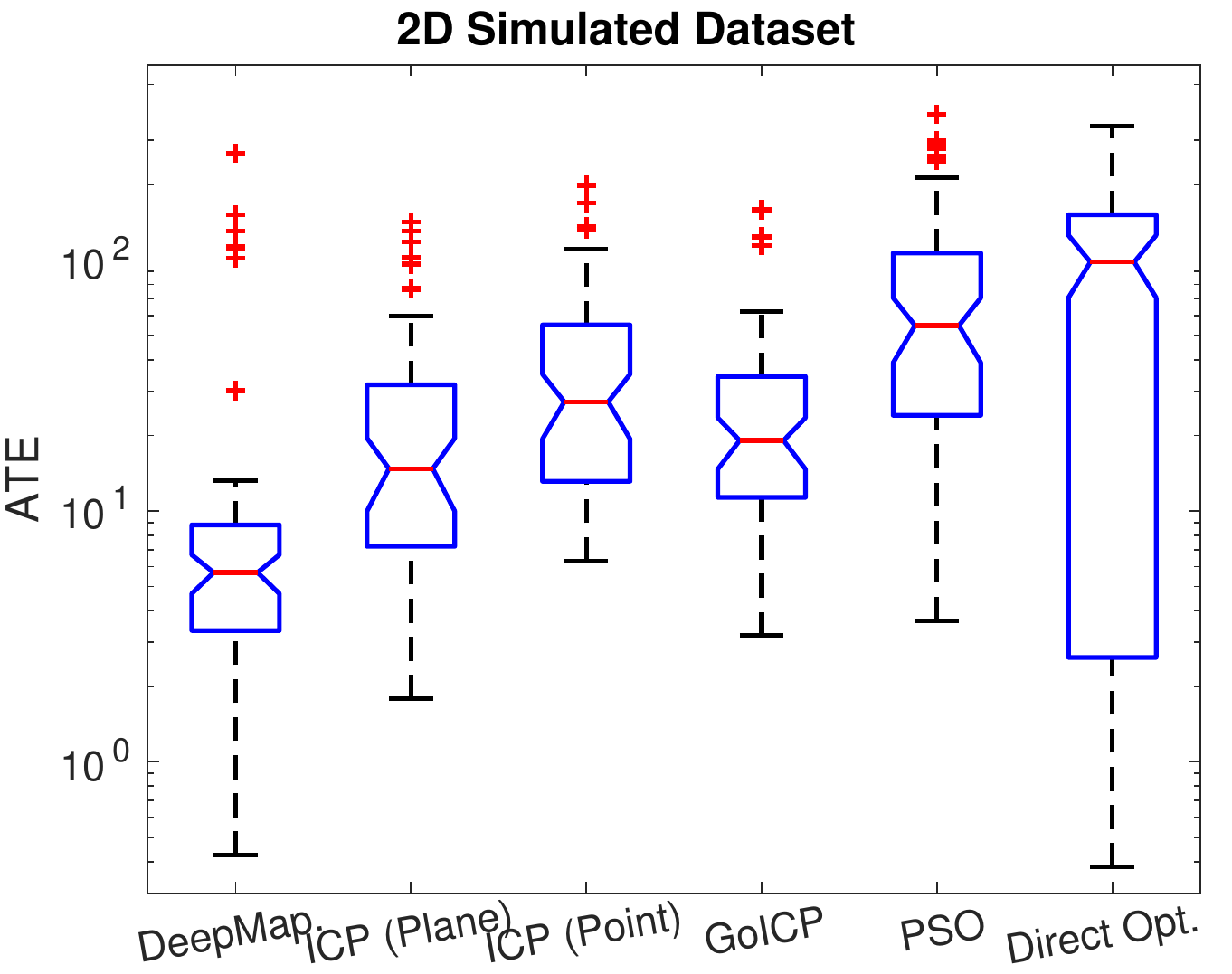}
  \centerline{(a)}
\end{minipage}
\hfill
\begin{minipage}[b]{0.49\linewidth}
  \centering
  \includegraphics[width=1\linewidth]{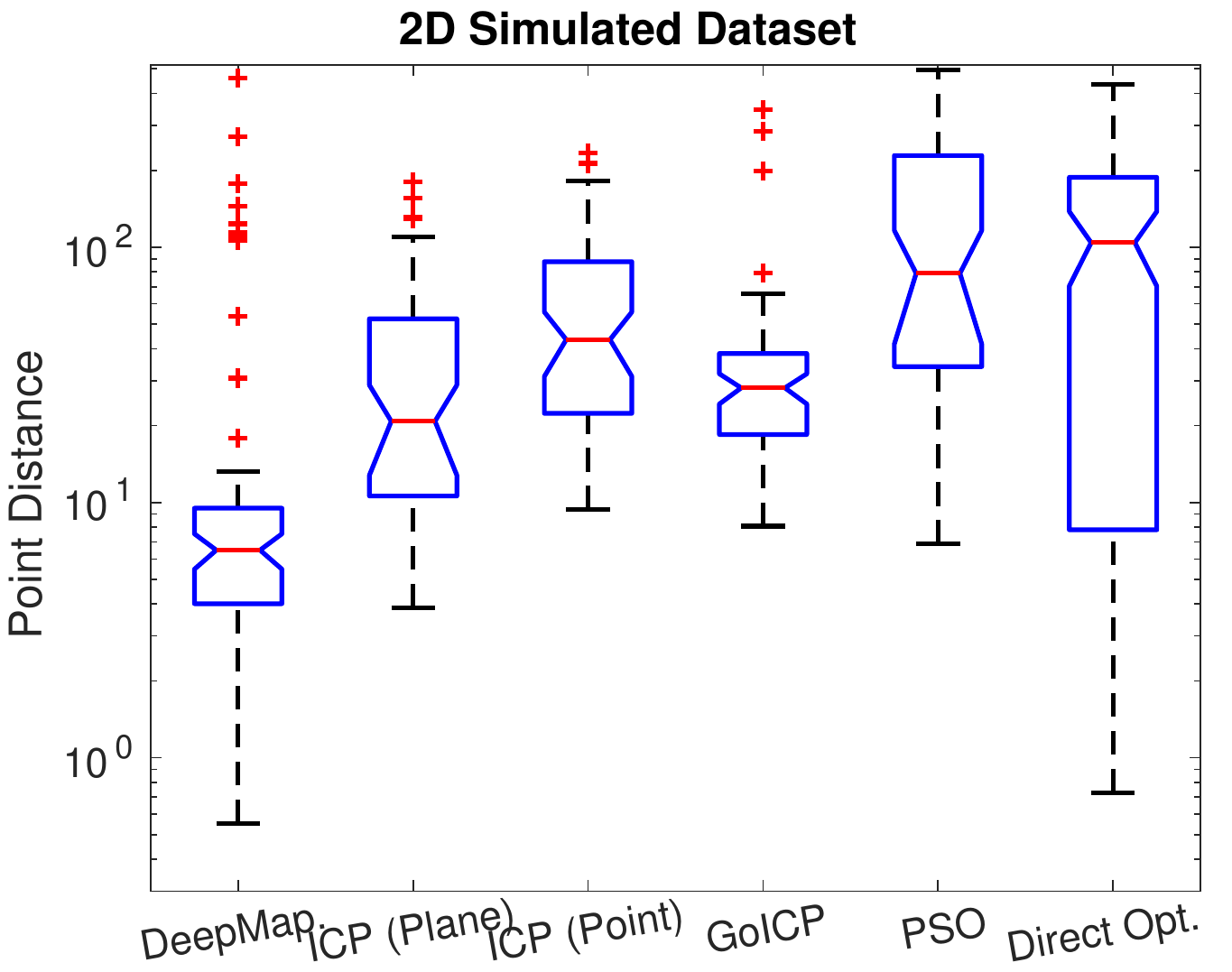}
  \centerline{(b)}
\end{minipage}
\caption{Box plots of the ATE and the point distance on the 2D simulated dataset. In each box, the red line indicates the median. The top and bottom blue edges of the box show the first (25th percentile) and the third (75th percentile) quartiles, respectively. Note the logarithmic scale for the y-axis.}
\label{fig:2d_quant} 
\end{figure}

The box plots in Figure~\ref{fig:2d_quant} show the quantitative results of the ATE and the point distance. Note that the data is plotted with the logarithmic scale for the y-axis. DeepMapping has the best performance in both metrics, achieving a median ATE of 5.7 pixels and a median point distance of 6.5, which significantly outperforms the baseline methods.

\begin{figure}[!h]
\centering
\begin{minipage}[b]{0.49\linewidth}
\centering
\includegraphics[width=1\linewidth]{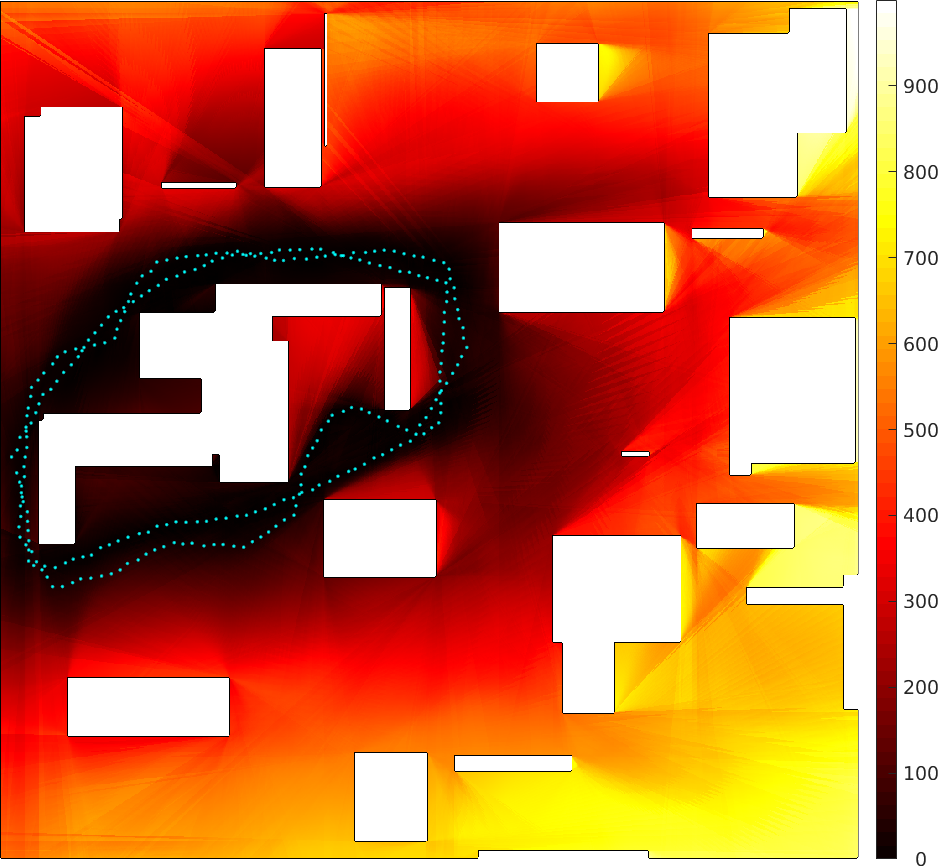}
\end{minipage}
\hfill
\begin{minipage}[b]{0.49\linewidth}
\centering
\includegraphics[width=1\linewidth]{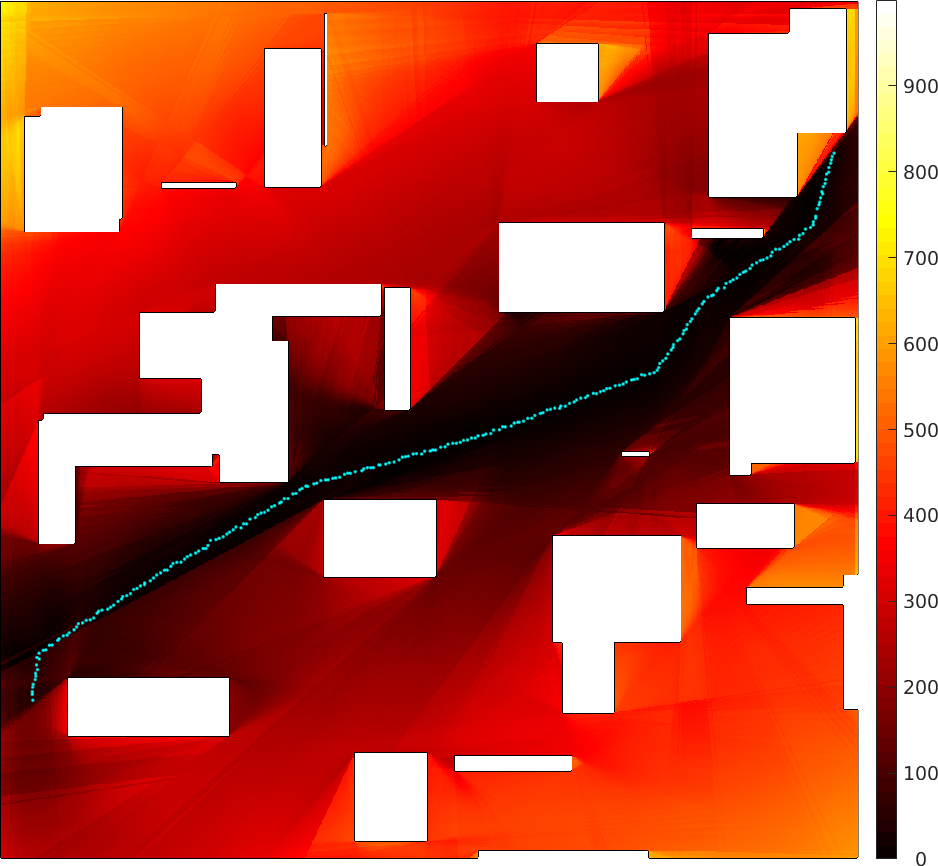}
\end{minipage}
\caption{Re-localization errors of two L-Nets (black is better). Cyan points show the trajectories used in the ``training" stage. Best viewed in color. \label{fig:position_error}}
\end{figure}

{\bf Generalization of L-Net:} while we do not expect the trained DeepMapping to generalize to other scenes, it is worth emphasizing that, to some extent, the ``trained" L-Net is able to perform coarse re-localization for an unseen point cloud that is captured close to the ``training" trajectory and has a similar orientation. For demonstration, we simulate local point clouds at all possible positions (e.g., all white pixels in Figure~\ref{fig:2d_data}) and use the ``trained" L-Net to estimate the sensor location. Figure~\ref{fig:position_error} shows the re-localization errors of two L-Nets ``trained" on different trajectories, thus having different generalization abilities. Only the point clouds captured close to the ``training" trajectories have low errors. The ability of coarse re-localization without ``re-training" has potential usages in robotic applications, such as the kidnapped robot problem.

\subsection{Experiments on the Active Vision Dataset}
\vspace*{-1mm}
\label{ssec:exp_avd}
{\bf Dataset: } DeepMapping is also tested on a real 3D dataset: Active Vision Dataset~\cite{Ammirato_AVD_ICRA17} that provides RGB-D images for a variety of scans of the indoor environment. Unlike other RGB-D datasets~\cite{Sturm_TUM_IROS12,Song_Sun3D_CVPR15} that capture RGB-D videos around the scene, the AVD uses a robotic platform to visit a set of discrete points on a rectangular grid with a fixed width of $300$mm. At each point, $12$ views of images are captured by rotating the camera every $30\degree$. This data collection procedure leads to a low frame rate and relatively small overlapping between two images.

The ground truth camera intrinsic and extrinsic parameters are provided. We use the camera intrinsic parameters to covert the depth map to point cloud representation and do not use any information provided by color images. We randomly move or rotate the camera sensor in the space and collect $105$ trajectories from the dataset. Each trajectory contains either $8$ or $16$ point clouds. 
\begin{figure*}[t]
\centering
\includegraphics[width=1\linewidth]{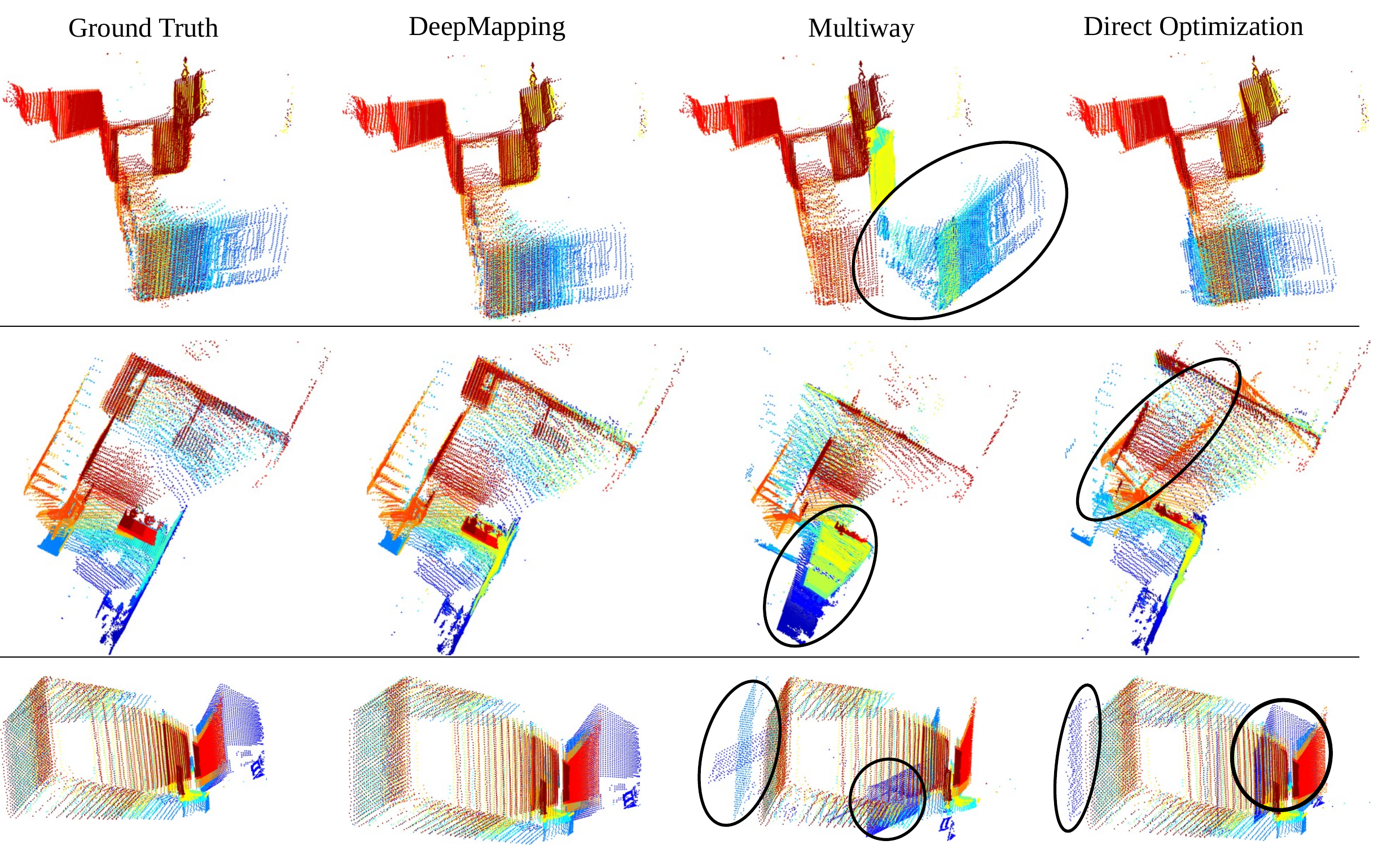}
\caption{Qualitative results of multiple point clouds registration tested on the AVD~\cite{Ammirato_AVD_ICRA17} dataset. The black ellipses highlight the misaligned parts in baselines. Each color represents one point cloud. More visualizations are shown in supplementary material. Best viewed in color.}
\label{fig:avd_qual}
\end{figure*}

{\bf Baseline:} we compare DeepMapping with the baseline method of multiway registration~\cite{Choi_ReconIndoor_CVPR15}. The ICP algorithms perform poorly because of the low overlapping rate between consecutive point clouds and thus are not included in this section. The multiway registration aligns multiple point clouds via pose graph optimization where each node represents an input point cloud and the graph edges connect two point clouds. We follow the same procedure in Section~\ref{ssec:exp_simulated} to test the performance of direct optimization with respect to sensor pose on the AVD. For DeepMapping and the direct optimization, we compare two loss functions: only the BCE loss $\mathcal{L}_{bce}$ (by setting $\lambda$ to $0$), and the combination of the BCE loss $\mathcal{L}_{bce}$ and the Chamfer distance  $\mathcal{L}_{ch}$. For the latter, we set $\lambda$ to $0.1$ because of the small overlapping between two scans.

{\bf Implementation: } the L-Net architecture consists of C$\lc 64 \rc$-C$\lc 128 \rc$-C$\lc 256 \rc$-C$\lc1024 \rc$-AM$\lc 1 \rc$-FC$\lc 512\rc$-FC$\lc 256\rc$-FC$\lc 3\rc$, where AM$\lc 1 \rc$ denotes 2D adaptive max-pooling layer. The M-Net has the same structure as that used in Section~\ref{ssec:exp_simulated}, except for the input layer that has 3 nodes rather than 2. We sample $35$ points on each laser ray. The multiway registration~\cite{Choi_ReconIndoor_CVPR15} is implemented using Open3D library~\cite{Zhou_Open3D_arxiv18}. The batch size is 8, and the optimization speed is approximately 125 epochs per minute on an Nvidia TITAN XP.

{\bf Results:}
The visual comparison of multiple point clouds registrations for different algorithms is shown in Figure~\ref{fig:avd_qual}. The misaligned point clouds are highlighted with black ellipses. The stairs from the direct optimization, for example, are not correctly registered, as shown in the second row in Figure~\ref{fig:avd_qual}. The multiway registration~\cite{Choi_ReconIndoor_CVPR15} and the direct optimization fails to align several planar structures that are shown in the last row in Figure~\ref{fig:avd_qual}. Failure cases of DeepMapping are shown in the supplementary material.

\begin{figure}[ht]
\begin{minipage}[b]{0.49\linewidth}
  \centering
  \includegraphics[width=1\linewidth]{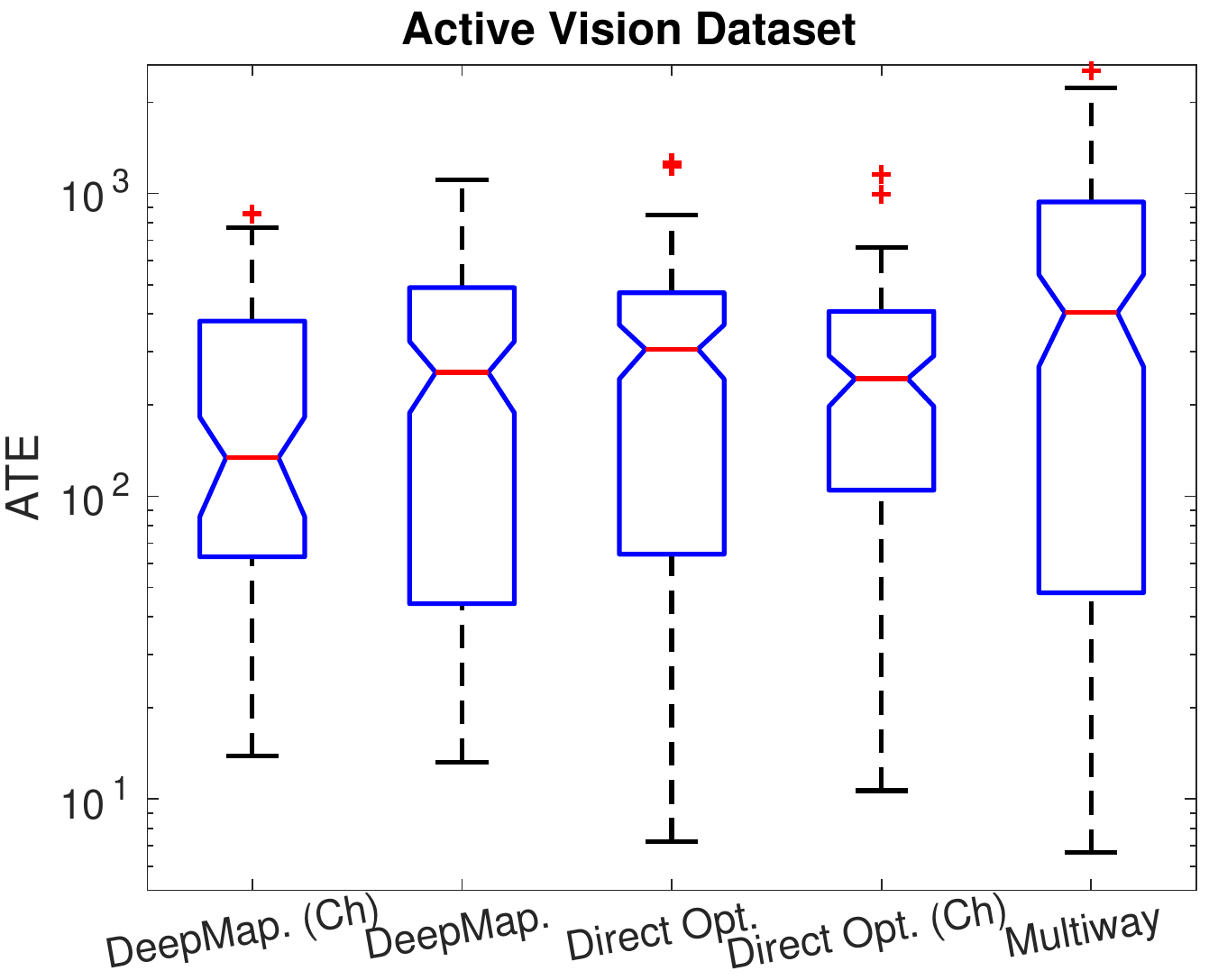}
  \centerline{(a)}
\end{minipage}
\hfill
\begin{minipage}[b]{0.49\linewidth}
  \centering
  \includegraphics[width=1\linewidth]{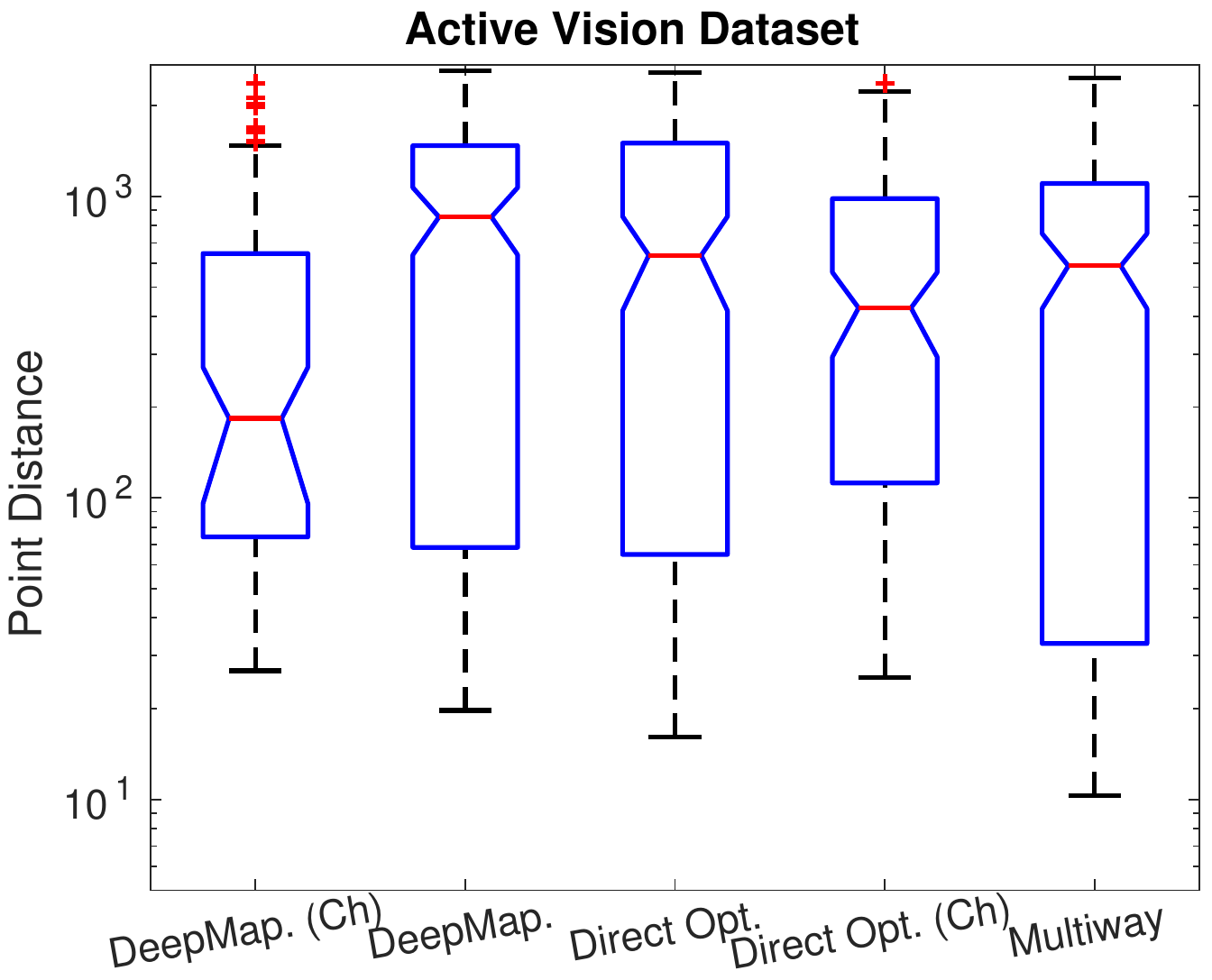}
  \centerline{(b)}
\end{minipage}
\caption{Box plots of the ATE and the point distance on the AVD~\cite{Ammirato_AVD_ICRA17}. The methods with ``Ch'' combine the Chamfer and the BCE losses. Legends are the same as those in Figure~\ref{fig:2d_qual}. Note the logarithmic scale for the y-axis.}
\label{fig:avd_quant} 
\end{figure}

We show the box plots of the ATE and the point distance in Figure~\ref{fig:avd_quant}. The multiway registration~\cite{Choi_ReconIndoor_CVPR15} performs poorly that has large errors in both metrics. Comparing the direct optimization with DeepMapping, we show the advantage of using neural networks for optimization. In addition, integrating Chamfer distance into DeepMapping is also helpful to improve the registration accuracy.

\section{Conclusion}
\label{sec:conclusion}
In this paper, we propose DeepMapping to explore a possible direction for integrating deep learning methods into multiple point clouds registration that could also be informative for other related geometric vision problems. The novelty of our approach lies in the formulation that converts solving the registration problem into ``training" some DNNs using properly defined unsupervised loss functions, with promising experimental performances.

\section*{Acknowledgment}
The authors gratefully acknowledge the helpful comments and suggestions from Yuichi Taguchi, Dong Tian, Weiyang Liu, and Alan Sullivan.  

\newpage
{\small
\bibliographystyle{ieee}
\bibliography{egbib}
}

\newpage
\appendix
\section*{Supplementary}

\section{Ablation Studies}
\label{sec:ablation}
In this section, we conduct several ablation studies to investigate the effects of various network architectures in DeepMapping using the AVD~\cite{Ammirato_AVD_ICRA17}. For quantitative comparison, we choose absolute trajectory error (ATE) as metrics and include the ATE from baseline multiway registration method~\cite{Choi_ReconIndoor_CVPR15}.

{\bf Feature extraction module in the L-Net:} we compare the effects of feature extraction module, i.e., CNN-based architecture and PointNet-based architecture~\cite{Qi_PointNet_CVPR17}. The CNN-based network consists of C$\lc 64 \rc$-C$\lc 128 \rc$-C$\lc 256 \rc$-C$\lc1024 \rc$-AM$\lc 1 \rc$, where C$\lc n\rc$ denotes 2D atrous convolutions that have kernel size $3$, dilation rate of $2$ and $n$-channel outputs,  AM$\lc 1 \rc$ denotes 2D adaptive max-pooling layer.. The PointNet-based architecture is FC$\lc 64 \rc$-FC$\lc 128 \rc$-FC$\lc 256 \rc$-FC$\lc1024 \rc$-AM$\lc 1 \rc$ , where FC$\lc n \rc$ denotes fully-connected layer with $n$-channel output.

\begin{figure}[b]
\centering
\includegraphics[width=0.5\linewidth]{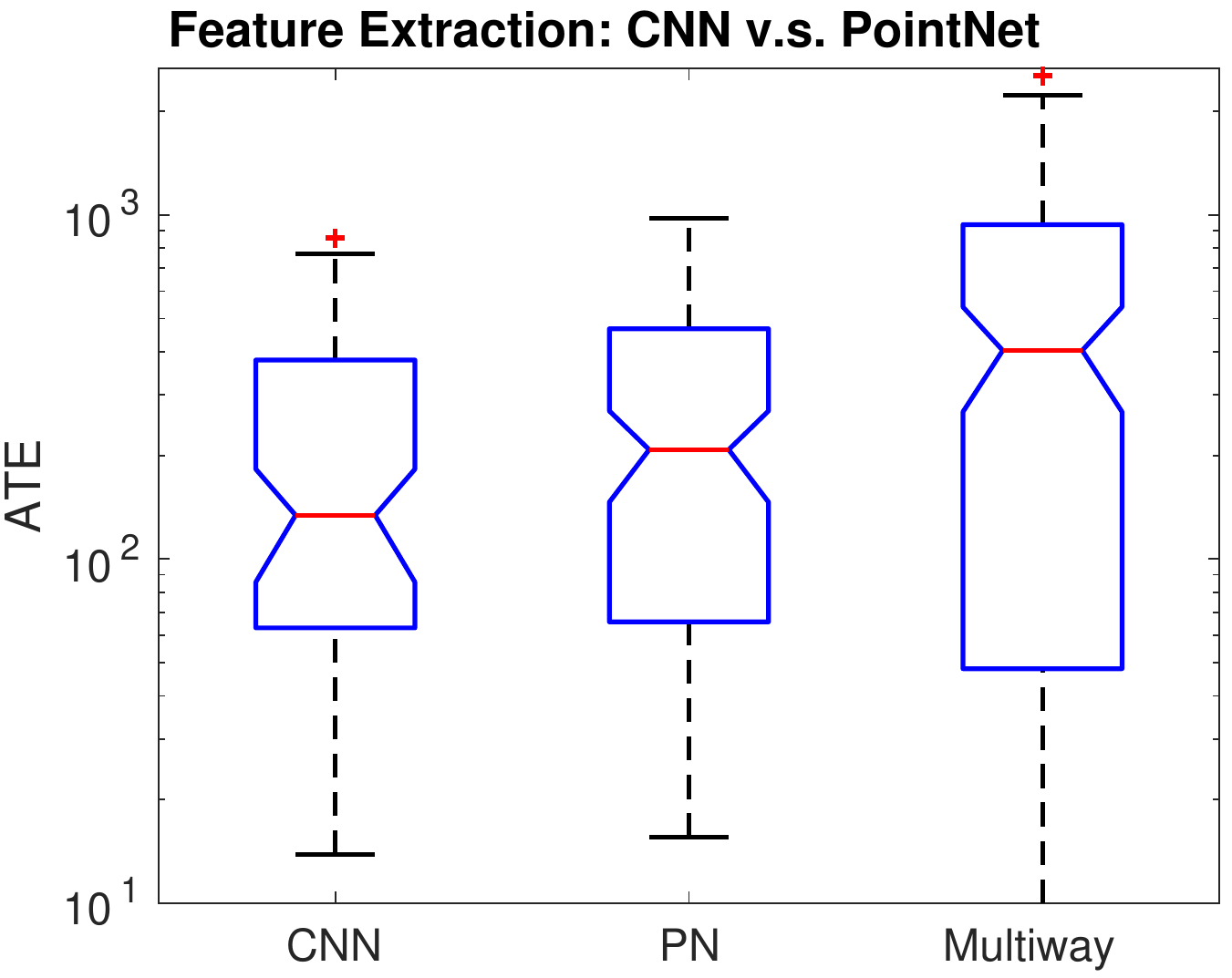}
\caption{Quantitative comparison of the ATE between CNN-based and PointNet-based architectures in the L-Net, tested on the AVD~\cite{Ammirato_AVD_ICRA17}.}
\label{fig:ablation_avd_cnn_pn}
\end{figure}

The box plot in Figure~\ref{fig:ablation_avd_cnn_pn} depicts the quantitative results of the ATE. As shown, CNN-based architecture achieves better performance with a median error of $134.07$mm than PointNet-based architecture that has a median error of $207.84$mm. This is not supervising because CNN is able to explore local structure information from neighborhood pixels while PointNet is a per-point function performing on each point independently.

\begin{figure}[hb]
\centering
\includegraphics[width=0.5\linewidth]{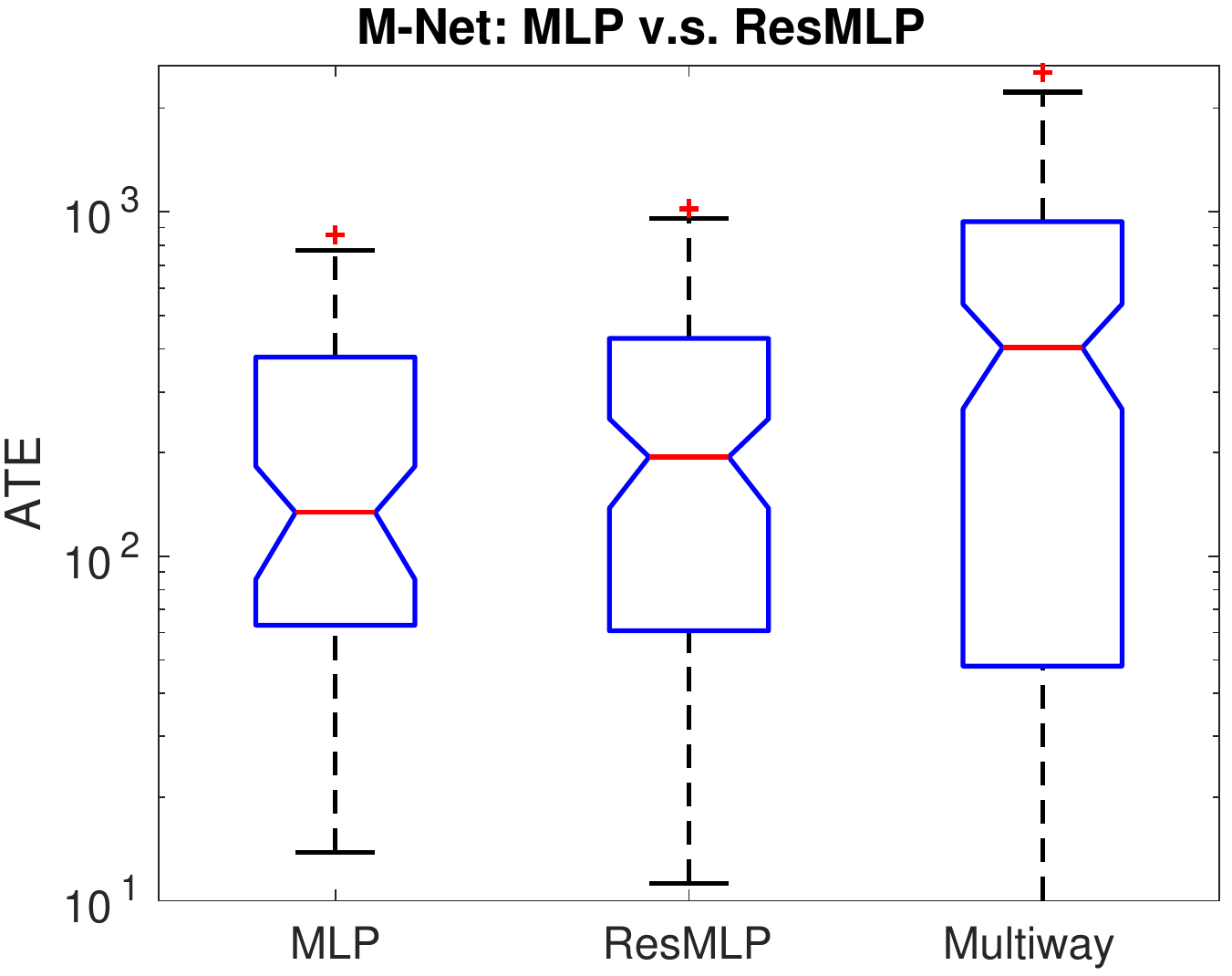}
\caption{Quantitative comparison of the ATE between MLP and ResMLP in the M-Net, tested on the AVD~\cite{Ammirato_AVD_ICRA17}.}
\label{fig:ablation_avd_mlp_resmlp}
\end{figure}

{\bf Architecture of the M-Net:} the proposed DeepMapping uses MLP in the M-Net to predict the occupancy status in the global coordinates. We compare this architecture with ResMLP that integrate the idea of deep residual networks~\cite{He_ResNet_CVPR16}. ResMLP consists of a stack of basic residual blocks where each residual block, denoted as RB$\lc n \rc$, contains two fully-connected layers with the same number of output nodes $n$. The detailed ResMLP architecture can be described as RB$\lc 64\rc$-RB$\lc 64\rc$-RB$\lc 64\rc$-RB$\lc 128\rc$-RB$\lc 128\rc$. As shown in Figure~\ref{fig:ablation_avd_mlp_resmlp}, MLP has a marginal improvement over ResMLP in terms of the ATE and therefore is adopted in the proposed DeepMapping.

\begin{figure}[tb]
\centering
\includegraphics[width=0.5\linewidth]{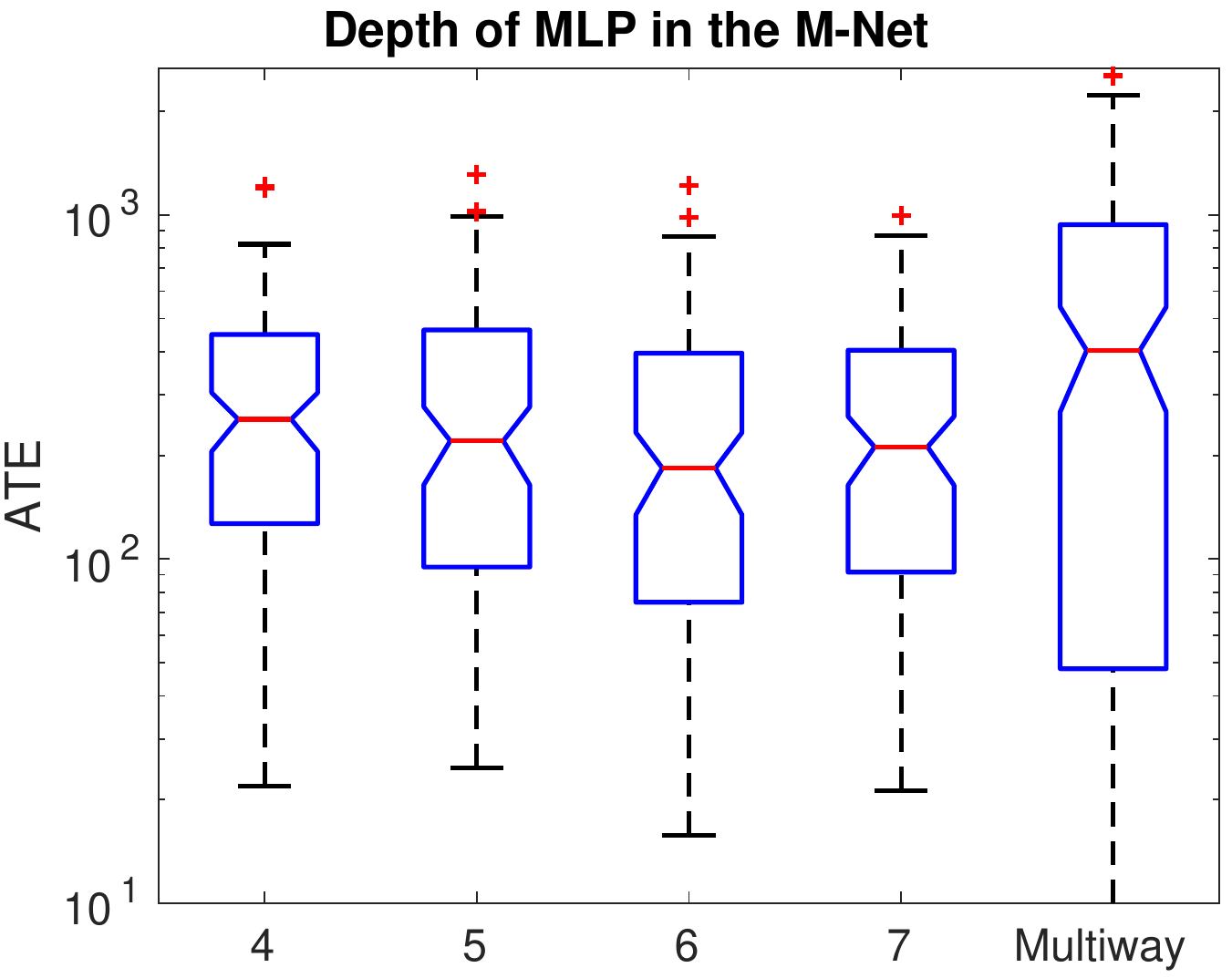}
\caption{Quantitative comparison of different depths of MLP in the M-Net, tested on the AVD~\cite{Ammirato_AVD_ICRA17}. The layer width is fixed to $64$.}
\label{fig:ablation_avd_mlp_depth}
\end{figure}

\begin{figure}[tb]
\centering
\includegraphics[width=0.5\linewidth]{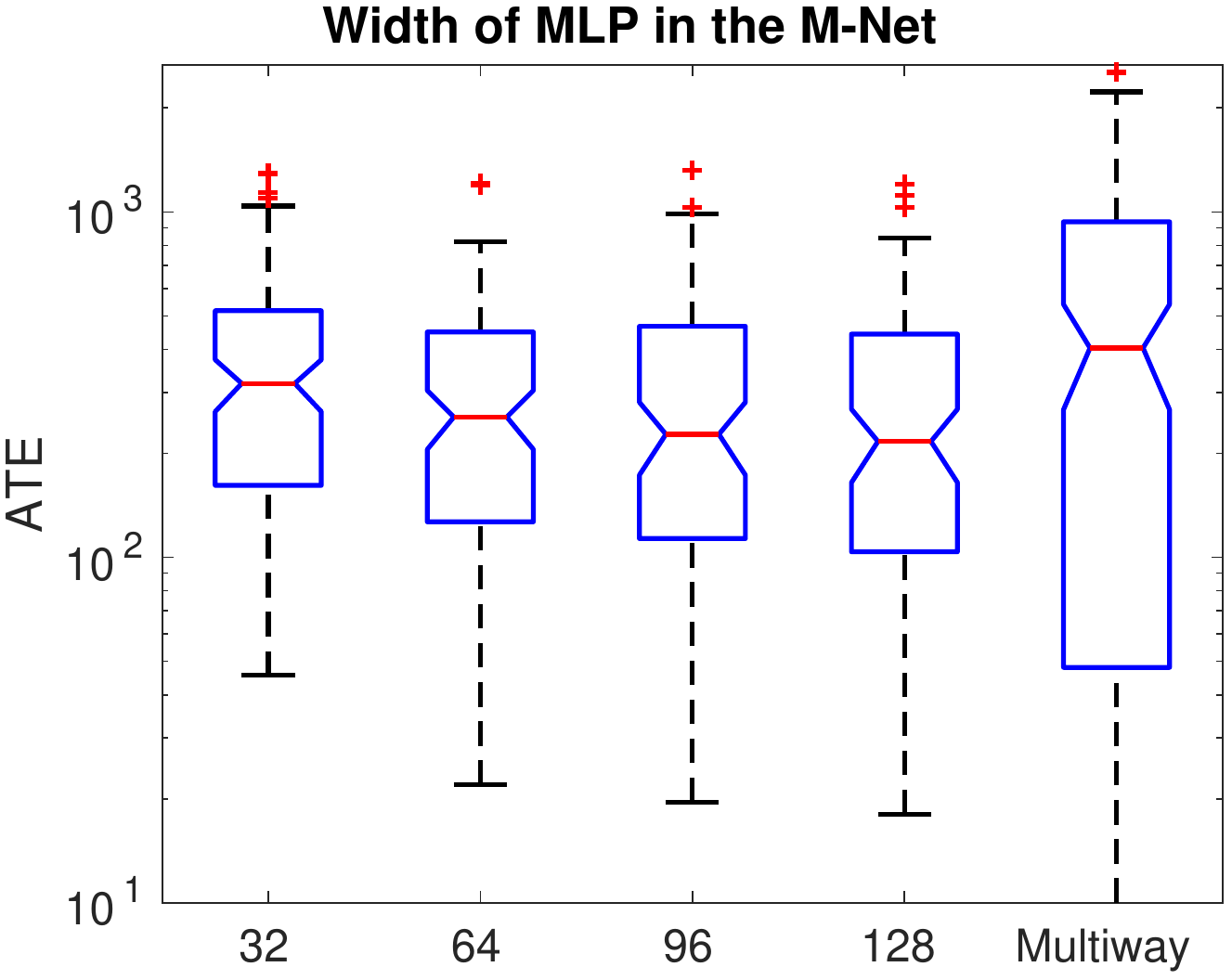}
\caption{Quantitative comparison of different width of MLP in the M-Net, tested on the AVD~\cite{Ammirato_AVD_ICRA17}. All MLP have the same depth of $4$ layers.}
\label{fig:ablation_avd_mlp_width}
\end{figure}

{\bf Depth and width of MLP in the M-Net:} the depth and width of MLP are defined as the number of layers in the MLP and the number of output nodes from each layer. To investigate the influence of layer depth, we fixed the layer width to $64$ and test MLP with depths $4$, $5$, $6$, and $7$. Figure~\ref{fig:ablation_avd_mlp_depth} show the corresponding results of the ATE. As shown, increasing MLP depth is beneficial to reducing the ATE. For example, MLP with depth $6$ has a lower error than those with depth $4$ and $5$. However, deeper networks may deteriorate the performance and make it difficult to optimize. To compare the effect of MLP width, we fixed the depth to $4$ and choose MLP with width $32$, $64$, $96$, and $128$. As shown in Figure~\ref{fig:ablation_avd_mlp_width}. MLP with a width of $128$ achieves the best performance. 

\section{More Results on 2D Simulated Point Cloud}
\label{sec:res_2D}
\begin{figure}[tb]
  \centering
  \includegraphics[height=0.6\columnwidth]{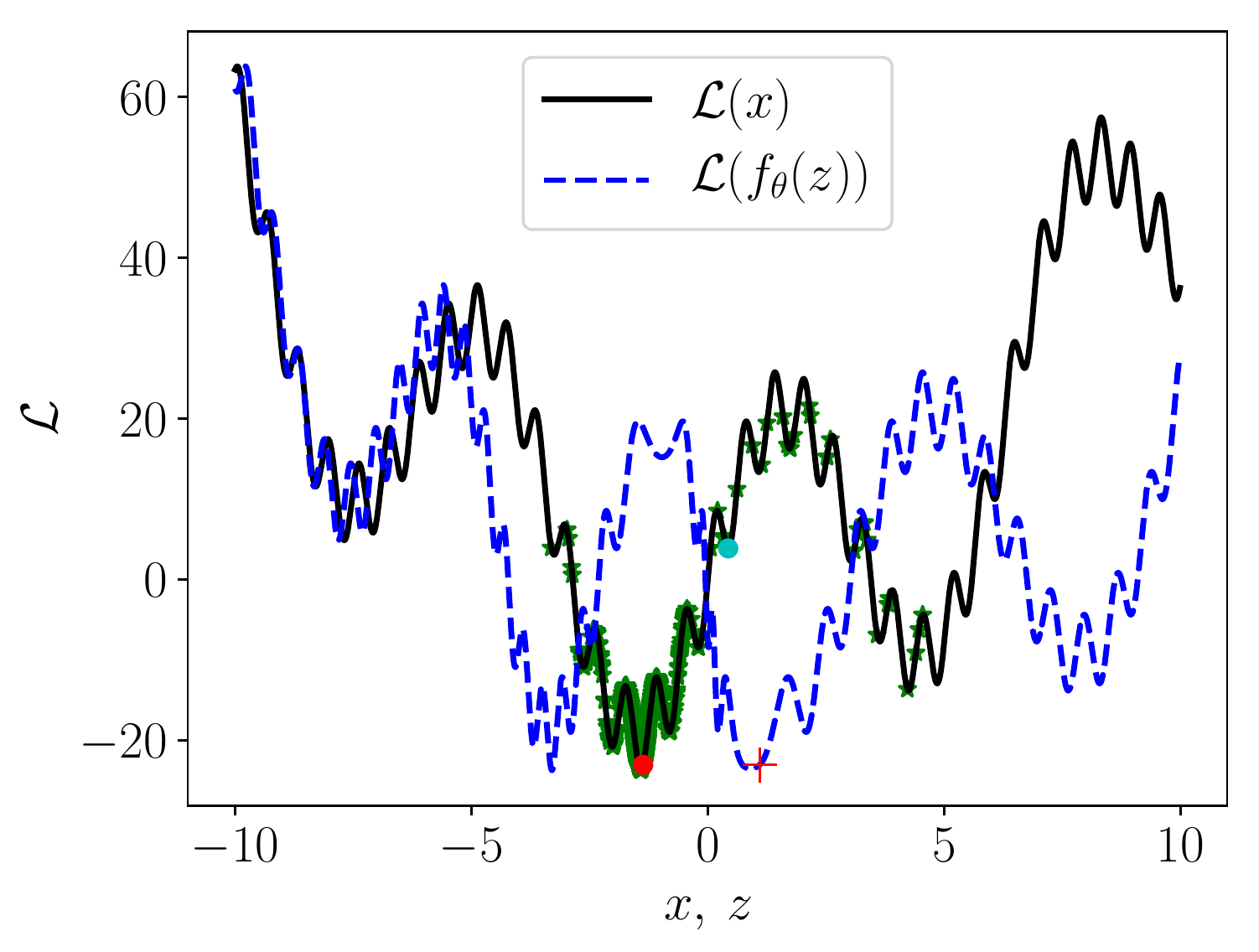}
  \caption{A 1D example to show the effectiveness of our neural-network-based conversion of a optimization problem into a higher dimension one. Red point shows the optimal solution found in the converted problem, while the cyan point shows the gradient descent optimum in the original problem. Please refer to Section~\ref{sec:interp} for a detailed explanation of the figure. Best viewed in color.}
  \label{fig:1d_ex} 
\end{figure}

\begin{figure*}[h]
\centering
\includegraphics[width=0.90\linewidth]{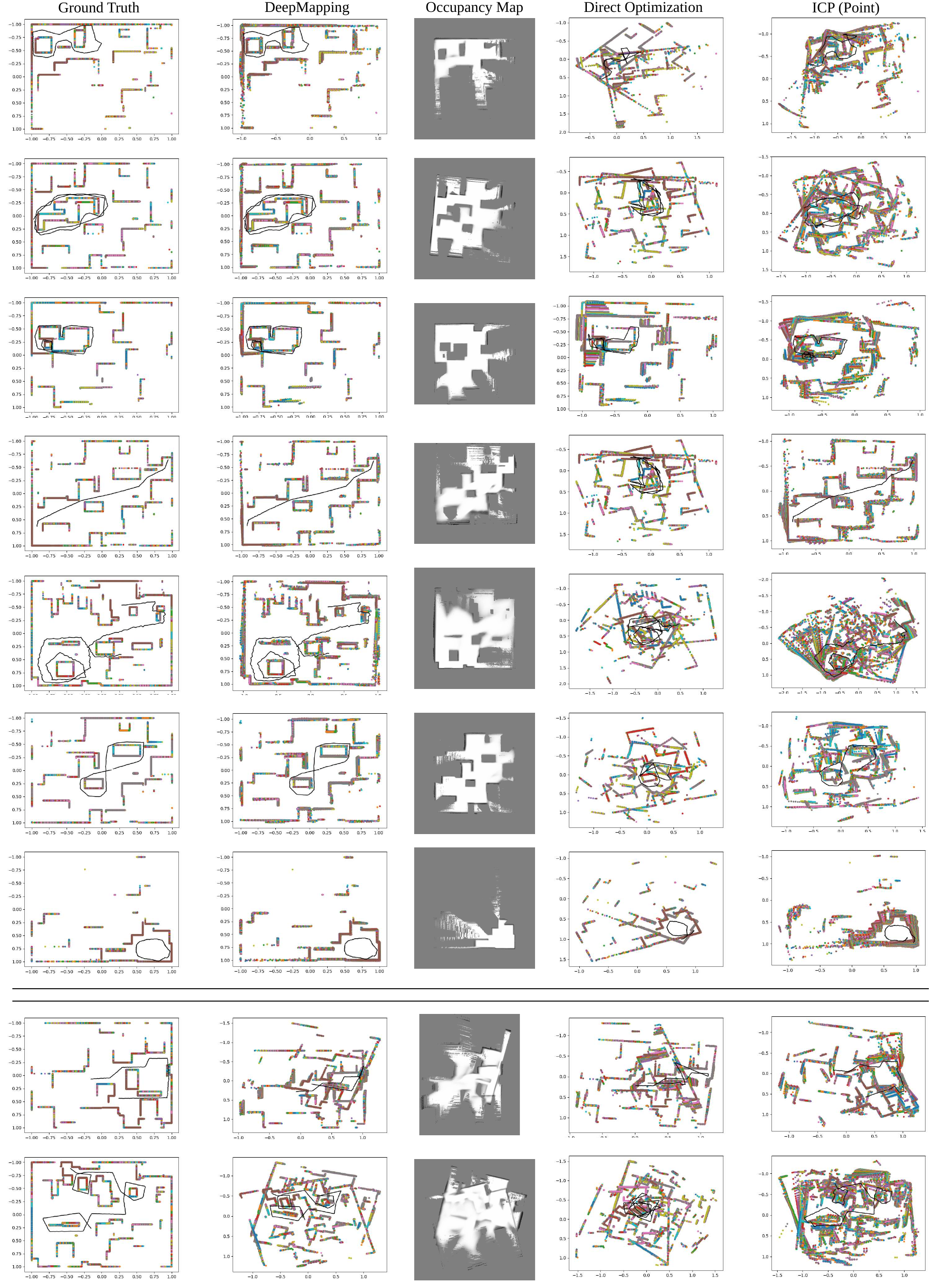}
\caption{Additional visual comparisons of multiple point clouds registration from the 2D simulated dataset. The black lines are the trajectories of sensor. The third column shows occupancy maps that are estimated by the M-Net. The black, while, and gray pixels show the occupied, unoccupied, and unexplored locations, respectively. Note that the results of each trajectory cam be defined in arbitrary coordinate systems and do not necessarily
aligned with ground truth. The last two rows show the failure cases. Best viewed in color.}
\label{fig:2d_qual_suppl}
\end{figure*}

Figure~\ref{fig:2d_qual_suppl} shows additional qualitative comparisons of registration results on the 2D simulated dataset. As shown, both the direct optimization and the incremental ICP with point-to-point metric fails to register all point clouds. The proposed DeepMapping, however, is more robust and accurate than baseline methods. The last two rows in Figure~\ref{fig:2d_qual_suppl} show two cases where all methods fail to find correct registration. 

Table~\ref{table:2d_time} reports the average execution time and the success rate for different methods to register $128$ point clouds. We run DeepMapping and the direct optimization for 3000 epochs. A registration of multiple point clouds is considered successful if the ATE is less than a threshold of 20 pixel, which is about $2\%$ of the image size ($1024\times 1024$). The success rate is then defined as the ratio of the number of successful registration to the total number of test cases. All methods are tested on a machine with a 3.3GHz Intel$^\circledR$ Core\texttrademark~i9-7900X CPU. We use an Nvidia TITAN XP for ``training" DeepMapping and the direct optimization. While DeepMapping seems slow, the method in fact converges very quickly: within 500 epochs (4.8 minutes), our ATE error is already smaller than of baseline methods.
\begin{table}[ht]
\resizebox{\columnwidth}{!}{%
\begin{tabular}{|c|cccc|}
\hline
        & DeepMapping & Direct Opt. & ICP (Point) & ICP (Plane) \\ \hline
Runtime & 29min      & 17min               & 6.48s       & 12.35s      \\ \hline
Succ. Rate & $84.2\%$     & $31.5\%$               & $36.0\%$        & $53.3\%$    \\ \hline
\end{tabular}
}
\caption{Average runtime for 3000 epochs and success rate for different methods tested on the 2D simulated dataset.}
\label{table:2d_time}
\end{table}

We also test two initialization methods, random initialization and zero initialization, for the direct optimization. Both methods have worse performances than the initialization which is the same as DeepMapping.

\section{More Results on the Active Vision Dataset}
\label{sec:res_avd}
\begin{figure*}[h]
\centering
\includegraphics[width=0.88\linewidth]{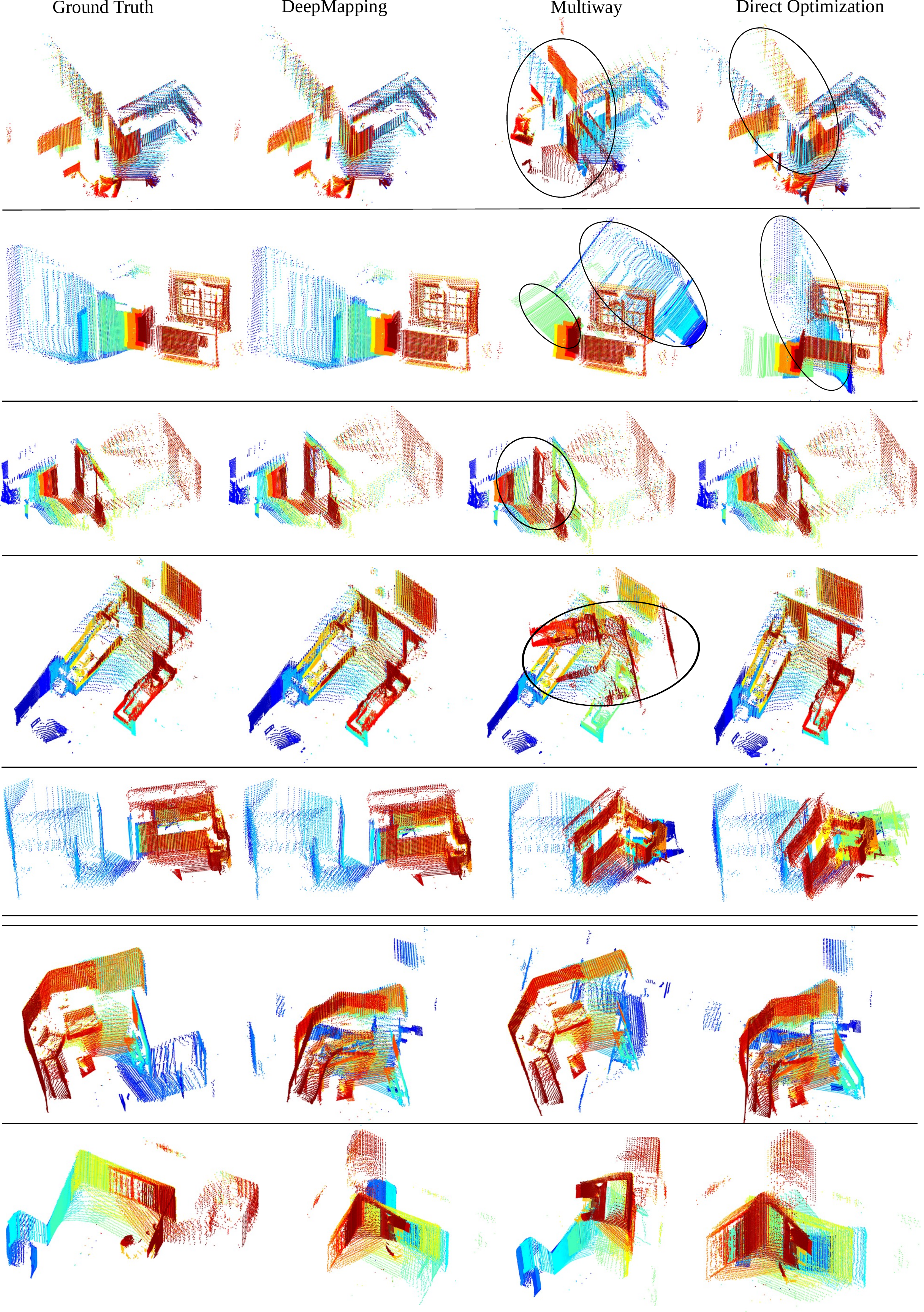}
\caption{Additional visual comparisons from the AVD~\cite{Ammirato_AVD_ICRA17}. The black ellipses highlight the misaligned parts in baselines. Each color represents one point cloud. The last two rows show the failure cases. Best viewed in color.}
\label{fig:avd_qual_suppl}
\end{figure*}

Figure~\ref{fig:avd_qual_suppl} shows additional visual comparison tested on the AVD~\cite{Ammirato_AVD_ICRA17}. The black ellipse highlights the region corresponding to misaligned parts from baseline methods. Table~\ref{table:avd_time} lists the average execution time for 3000 epochs and the success rate to register $16$ point clouds from the AVD. In this experiment, A registration is considered to be successful if the ATE is less than $450mm$. The hardware configuration is identical to those in Section~\ref{sec:res_2D}. As shown, the success rate from DeepMapping is higher than the rate from multiway registration~\cite{Choi_ReconIndoor_CVPR15}.

\begin{table}[ht]
\resizebox{\columnwidth}{!}{%
\begin{tabular}{|c|ccc|}
\hline
        & DeepMapping & Direct Opt. & Multiway~\cite{Choi_ReconIndoor_CVPR15} \\ \hline
Runtime & 24min      & 20min               & 42.49s       \\ \hline
Succ. Rate & $80.0\%$      & $77.1\%$               & $58.1\%$           \\ \hline
\end{tabular}
}
\caption{Average runtime for 3000 epochs and success rate for different methods tested on the AVD.}
\label{table:avd_time}
\end{table}

\section{Interpretation of Our Method}
\label{sec:interp}

Given the differences between the problem formulations in~\eqref{eq:trad_loss} and~\eqref{eq:prop_loss}, it is natural to ask why we use the neural network $f_\theta$ to estimate the sensor poses $\mT$ instead of directly optimizing them. In this section, we attempt to provide a simple potential interpretation of the benefit introduced by our formulation.

The basic inspiration comes from an optimization technique known as changing variables~\cite{Agrawal_ChgVar_JCD18} that can convert an originally non-convex optimization problem to an equivalent convex one. In their example, a geometric program can be converted to a linear program by substituting exponential functions as original variables. In our formulation, we combine this idea with neural networks by replacing the optimization variables $\mT$ with $f_\theta\lc \mS\rc$ and transforming the objective function from~\eqref{eq:trad_loss} to~\eqref{eq:prop_loss}. While we do not expect that the replacement of variables $\mT$ with neural network parameters $\theta$ yields a convex problem, we observe that this transformation is beneficial to finding the optimal solution to the original problem. 

We conduct a simple 1D experiment to illustrate this observation. Consider a problem of finding the optimal value of $x\in \mathbb{R}$ that minimizes $\mathcal{L}\lc x\rc$, a non-convex objective function with multiple local minima shown as the black line in Figure~\ref{fig:1d_ex}. Specifically, the objective function is defined as 
\begin{equation*}
\mathcal{L}\lc x \rc = \frac{1}{2} x^2 + 5\sin\lc 10x \rc + 20\sin \lc x \rc \textrm{.}
\label{eq:obj_func}
\end{equation*}

In this experiment, we compare two optimization methods, i.e., the proposed network-based optimization and the direct optimization. For network-based optimization, we introduce an MLP, $f_\theta$, which consists of FC$\lc 10\rc$-FC$\lc 20\rc$-FC$\lc 30\rc$-FC$\lc 40\rc$-FC$\lc 1\rc$. Each MLP layer is followed by an ELU~\cite{Clevert_ELUs_ICLR15} activation function except for the output layer. The MLP has one node in the input and the output layer to replace the variable $x$ with $f_\theta\lc z\rc$, resulting in another problem with optimization variable $z$. To ensure the same starting point, the direct optimization is initialized with $x_0 = f_{\theta_0} \lc z_0 \rc$ where $\theta_0$ and $z_0$ are the initial values of network parameters $\theta$ and variable $z$, respectively. We use gradient descent with a learning rate of $2\times 10^{-4}$ and run $1000$ iterations. For the network-based optimization, we jointly update the network parameters $\theta$ and $z$.

The cyan point shows the result using gradient descent optimization that is performed directly on $\mathcal{L}\lc x\rc$, which is trapped in a local minimum. The function $\mathcal{L}\lc f_{\theta{^\star}} \lc z \rc \rc$ with the optimal $\theta^\star$ found in the network-based optimization is plotted as the blue dash line in Figure~\ref{fig:1d_ex}. We take $f_{\theta{^\star}} \lc z^\star \rc$ to retrieve the optimal point $x^\star$ for $\mathcal{L}\lc x\rc$. The red plus and red circle in Figure~\ref{fig:1d_ex} correspond to $z^\star$ and $x^\star$, respectively. The green star symbols show the values of $x$ during the 1000 gradient-descent iterations.

Notice the distribution of the green star symbols, visualizing the ``sampled locations'' in the domain, $x$, of the original problem. It is interesting to see that our conversion leads to a wider search range in the original problem domain, while keeping the same number of function evaluations of the original problem $\mathcal{L}(\cdot)$ as in direct gradient descent.

\end{document}